\documentclass{article}
\usepackage{xspace}
\usepackage{microtype}
\usepackage{graphicx}
\usepackage{subcaption}
\usepackage{booktabs} 

\usepackage{subcaption}
\usepackage{mathtools}
\usepackage{adjustbox}
\usepackage{xcolor} 
\usepackage{colortbl}
\usepackage{multirow}

\newcommand{\etal}{\textit{et~al.}\xspace}

\usepackage{hyperref}

\usepackage[preprint]{icml2026}

\usepackage{amsmath}
\usepackage{amssymb}
\usepackage{mathtools}
\usepackage{amsthm}

\usepackage[capitalize,noabbrev]{cleveref}

\theoremstyle{plain}

\theoremstyle{definition}

\theoremstyle{remark}

\usepackage[textsize=tiny]{todonotes}

\begin{document}

\twocolumn[
  
  \icmltitle{Learning Hierarchical Sparse Transform Coding for 3DGS Compression}



  \icmlsetsymbol{equal}{*}

  \begin{icmlauthorlist}
    \icmlauthor{Hao Xu}{yyy}
    \icmlauthor{Xiaolin Wu}{sch}
    \icmlauthor{Xi Zhang}{comp}
  \end{icmlauthorlist}

  \icmlaffiliation{yyy}{McMaster University, Canada}
  \icmlaffiliation{comp}{Nanyang Technological University, Singapore}
  \icmlaffiliation{sch}{Southwest Jiaotong University, China}

  \icmlcorrespondingauthor{Xiaolin Wu}{xlw@swjtu.edu.cn}

  \icmlkeywords{Machine Learning, ICML}

  \vskip 0.3in

]



\printAffiliationsAndNotice{}  

\begin{abstract}
Current 3DGS compression methods largely forego the neural analysis-synthesis transform, which is a crucial component in learned signal compression systems. As a result, redundancy removal is left solely to the entropy coder, overburdening the entropy coding module and reducing rate-distortion (R-D) performance. To fix this critical omission, we propose a training-time transform coding (TTC) method that adds the analysis-synthesis transform and optimizes it jointly with the 3DGS representation and entropy model. Concretely, we adopt a hierarchical design: a channel-wise KLT for decorrelation and energy compaction, followed by a sparsity-aware neural transform that reconstructs the KLT residuals with minimal parameter and computational overhead. Experiments show that our method delivers strong R–D performance with fast decoding, offering a favorable BD-rate–decoding-time trade-off over SOTA 3DGS compressors.
\end{abstract}

\begin{figure*}[htb]
  \begin{center}
    \centerline{\includegraphics[width=\textwidth]{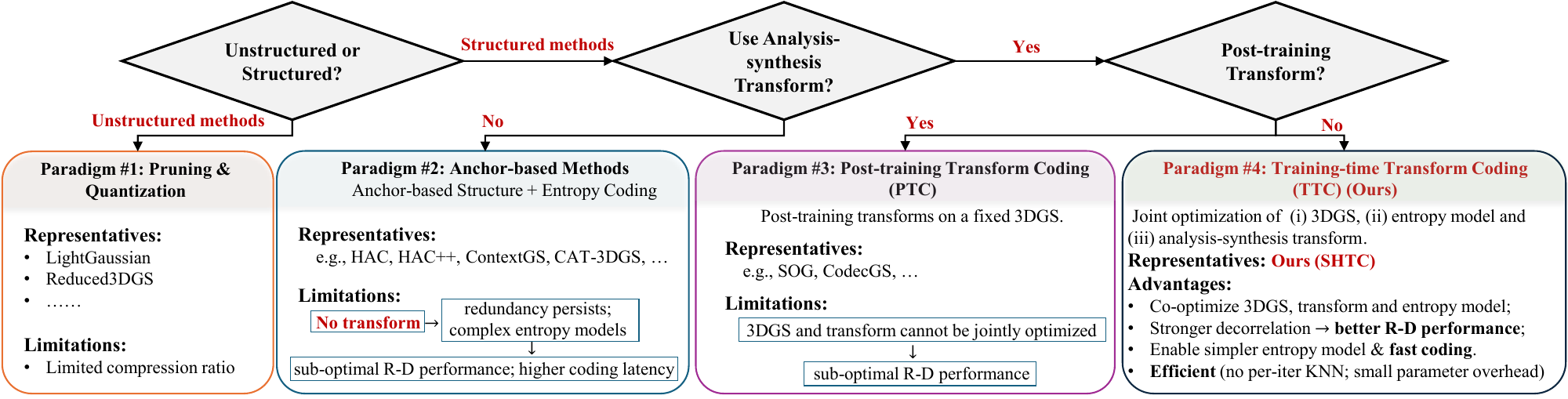}}
    \caption{
      Overview of 3DGS compression paradigms.
 Methods are categorized into four paradigms: (1) pruning and quantization, (2) anchor-based structure with adaptive entropy coding, (3) post-training transform coding (PTC) on a frozen 3DGS, and (4) training-time transform coding (TTC), introduced by us, which is the first to jointly optimize 3DGS, context model and the transform for improved R-D performance. Within TTC, we further propose SHTC, a lightweight transform architecture that delivers strong R-D performance with minimal parameter and computation overhead.
    }
    \label{fig:Teaser}
  \end{center}
\end{figure*}

\section{Introduction}
3D Gaussian Splatting (3DGS)~\cite{kerbl20233d} enables real-time, high-quality novel-view synthesis, but its many Gaussian primitives create significant storage and bandwidth overheads, motivating 3DGS compression.
As summarized in Fig.~\ref{fig:Teaser}, existing 3DGS compression methods can be broadly grouped into unstructured approaches (e.g., pruning and quantization) and structured, entropy-coded pipelines. Within the structured category, most competitive compressors either adopt the anchor-based paradigm by pairing Scaffold-GS~\cite{lu2024scaffold} with context-based entropy coding, or perform post-training transform coding (PTC) on a fixed 3DGS. In anchor-based methods, end-to-end analysis–synthesis transform coding is absent, so redundancy is not sufficiently removed before entropy coding. This leaves high-dimensional dependencies and sparsity underexploited, shifting the burden to increasingly complex entropy models (e.g., HAC++~\cite{chen2025hac++}, ContextGS~\cite{wang2024contextgs}), which can still yield suboptimal rate-distortion (R–D) performance and higher decoding latency.
Although a few prior works recognize the importance of transforms~\cite{morgenstern2024compact,lee2025compression}, they apply them only post-hoc to a fixed  3DGS, decoupling transform from 3DGS representation learning and preventing mutual adaptation under a unified R–D objective, which often limits compression gains. To address these limitations, we advocate a new paradigm, training-time transform coding (TTC), in which analysis–synthesis transforms are learned during training, allowing the 3DGS representation, entropy model, and transforms to be jointly optimized under a unified R-D objective.

Unlike neural image/video compression where the transform is shared across images and incurs no per-image overhead, TTC uses scene-specific transforms that must be transmitted as part of the bitstream. From the minimum description length (MDL) perspective~\cite{rissanen1978modeling}, TTC should ideally balance the transform cost and the coding cost of the transformed 3DGS attributes, i.e., $L(M) + L(D|M)$, where $L(M)$ denotes the number of bits to describe the scene-specific transform and $L(D|M)$ is the code length for encoding the transformed 3DGS attributes. In principle, increasing transform complexity can reduce $L(D|M)$, 
but simultaneously increases $L(M)$, potentially canceling out the compression gain. 

Optimizing the full MDL objective for TTC is highly challenging, because it requires non-trivial discrete–continuous optimization over the transform’s parameters and architecture. MDL optimization problem is only well posed for a given  the model class. As the attainable trade-off between $L(M)$ and $L(D|M)$ is with respect to the chosen transform family, search-based solvers such as NAS~\cite{elsken2019neural} can only explore what this search space permits. Thus, a more important and unanswered question is how to design an appropriate transform architecture for scene-specific 3DGS compression. Once such an architecture defines a suitable transform family as the search space, future work can readily apply existing NAS to search for MDL-efficient designs. In this work, we focus on this prerequisite step and propose an MDL-aware transform architecture that is parameter- and computation-efficient.

To instantiate this MDL-aware, parameter- and computation-efficient transform design within TTC, we propose sparsity-guided hierarchical transform coding (SHTC). SHTC avoids computationally prohibitive repeated spatial KNN graph construction by restricting transforms to the channel domain, and uses a lightweight two-layer hierarchy to keep parameter and computational overhead low. In Layer 1, we apply the Karhunen-Lo\`eve Transform (KLT) to decorrelate channels and compact energy, yielding coefficients with highly uneven energy distribution. Coding only the principal coefficients reduces rate but introduces truncation error, whereas coding all coefficients is rate-inefficient. To balance this trade-off, Layer 2 serves as a refinement layer and uses a neural transform to compress the KLT residual, compensating for truncation-induced information loss with modest additional rate. Since the residual is typically compressible with many near-zero entries, we inject this sparsity prior as an inductive bias into the refinement layer, enabling effective residual coding with very few parameters. Concretely, we adopt a compressed-sensing-inspired refinement that encodes the residual with a small set of learned linear measurements and reconstructs it using a lightweight deep-unfolding decoder~\cite{gregor2010learning,zhang2020deep,zhang2018ista}.

\begin{figure}
  \begin{center}
    \centerline{\includegraphics[width=\columnwidth]{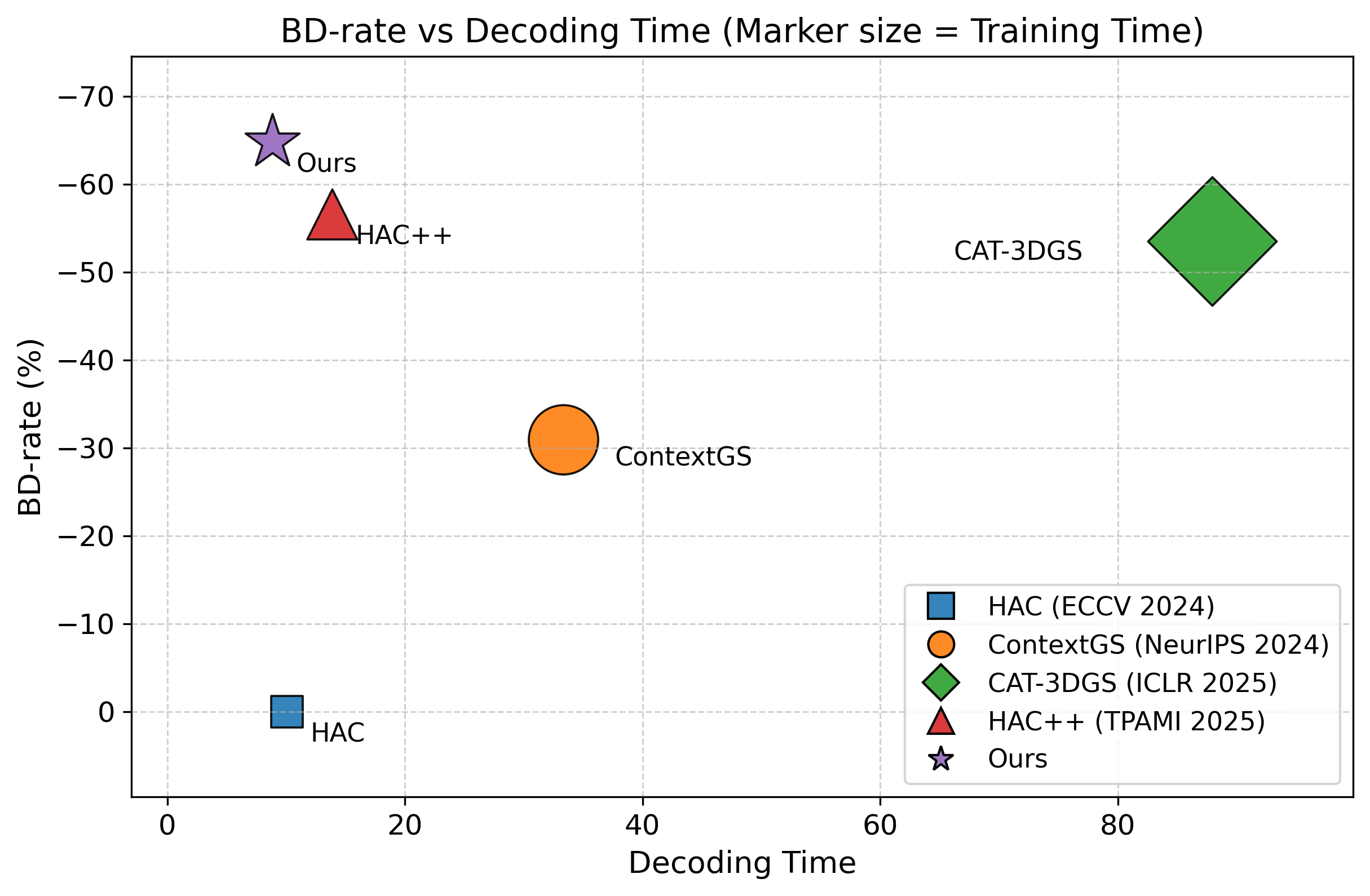}}
    \caption{
      Trade-off between decoding time and BD-rate (lower is better), with marker size indicating training time for HAC, ContextGS, CAT-3DGS, HAC++, and Ours.
    }
    \label{fig:pareto}
  \end{center}
\end{figure}

Overall, we introduce TTC, which enables joint optimization of the 3DGS representation, the entropy model, and the transforms. Within this paradigm, we propose SHTC, a parameter- and computation-efficient transform architecture that achieves substantial R–D gains with minimal decoding-time overhead and moderate training-time cost; moreover, Fig.~\ref{fig:pareto} shows that it is empirically Pareto-optimal in the BD-rate–decoding-time trade-off compared with current state-of-the-art 3DGS compressors. In addition, our parameter-efficient design strategy, which combines a sparsity prior with deep unfolding, may offer a potential blueprint for developing low-complexity neural image and video codecs.

\begin{figure*}[t]
    \centering
    \includegraphics[width=\linewidth]{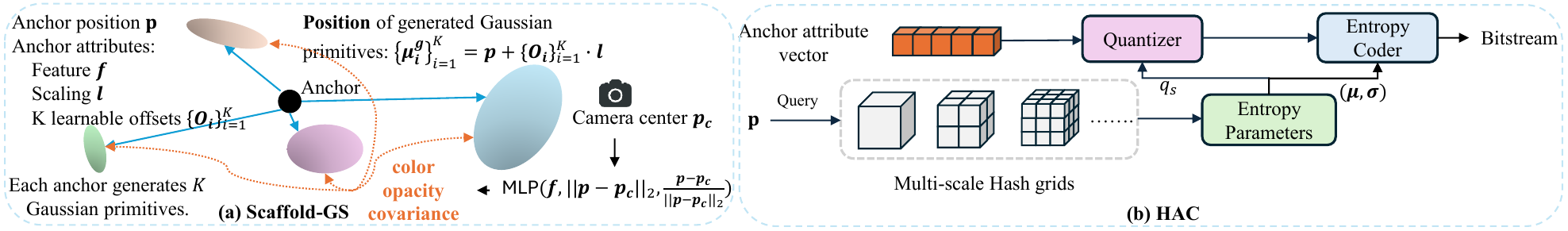}
    \caption{Illustration of (a) Scaffold-GS and (b) an anchor-based compression pipeline, using HAC as an example.}
    \label{fig:scaffold}
\end{figure*}
\section{Preliminary and Related Work}
\label{sec:scaffold}
In this work, to achieve the best compression performance, SHTC is integrated into anchor-based frameworks to enable the joint optimization of 3DGS, the entropy model, and the transforms. Since these anchor-based frameworks are built upon Scaffold-GS~\cite{lu2024scaffold}, we first recap the Scaffold-GS representation and then review how recent anchor-based methods further compress it. Due to space constraints, a more comprehensive introduction for the unstructured compression methods is deferred to Appendix~\ref{sec:add_related_work}; a dedicated review of post-training transform coding in existing 3DGS compression approaches is presented in Appendix~\ref{sec:3dgs_post_training_transform}. 

Scaffold-GS introduces a hierarchical structure during training: a sparse set of anchors serve as reference points from which a dense set of Gaussians are generated. As shown on the left side of Fig. \ref{fig:scaffold}, each anchor is parameterized by the position $\mathbf{p}$, latent feature $\mathbf{f}$, scaling factor $\mathbf{l}$, and $K$ learnable offsets $\{\mathbf{O}_i\}_{i=1}^{K}$. The scaling factor $\mathbf{l}$ together with the offsets $\{\mathbf{O}_i\}_{i=1}^{K}$ determine the spatial distribution of the generated Gaussians. For each generated Gaussian, the view-dependent color, opacity, and covariance (parameterized by quaternion + scale) are predicted from the anchor’s latent feature through lightweight MLPs. 

Although Scaffold-GS is more compact than vanilla 3DGS, it primarily performs representation compaction. Anchor parameters and attributes are still stored as raw 32-bit floats, without quantization or entropy coding, which fundamentally limits the achievable bitrate reduction. SOGS reduces the dimensionality of anchor features~\cite{zhang2025sogs}, but it still falls into the category of compaction rather than learned compression, and the gains remain limited.
To overcome these limitations, recent anchor-based methods introduce context-based conditional entropy models to quantize and entropy-code anchor attributes, substantially improving compression ratios. HAC is an early representative baseline in this line of work. As illustrated in the right panel of Fig.~\ref{fig:scaffold}, HAC employs multi-scale hash grids as a hyperprior~\cite{balle2018variational}. Given an anchor position, it queries the hash grids to obtain contextual features, which are then mapped by an MLP to the quantization step size $q_s$ and the Gaussian likelihood parameters $(\boldsymbol{\mu}, \boldsymbol{\sigma})$. Building on HAC, subsequent methods improve the rate--distortion performance by adopting more expressive entropy models, such as channel-wise autoregressive model (ChARM)~\cite{minnen2020channel} in HAC++~\cite{chen2025hac++} and CAT-3DGS~\cite{zhan2025catdgs}, as well as spatial autoregressive model (SARM) in ContextGS~\cite{wang2024contextgs} and HEMGS~\cite{liu2024hemgs}. These methods represent the current state-of-the-art compression performance for 3DGS compression.

\begin{figure*}[htbp]
    \centering
    \begin{subfigure}[b]{0.32\textwidth}
        \includegraphics[width=\linewidth]{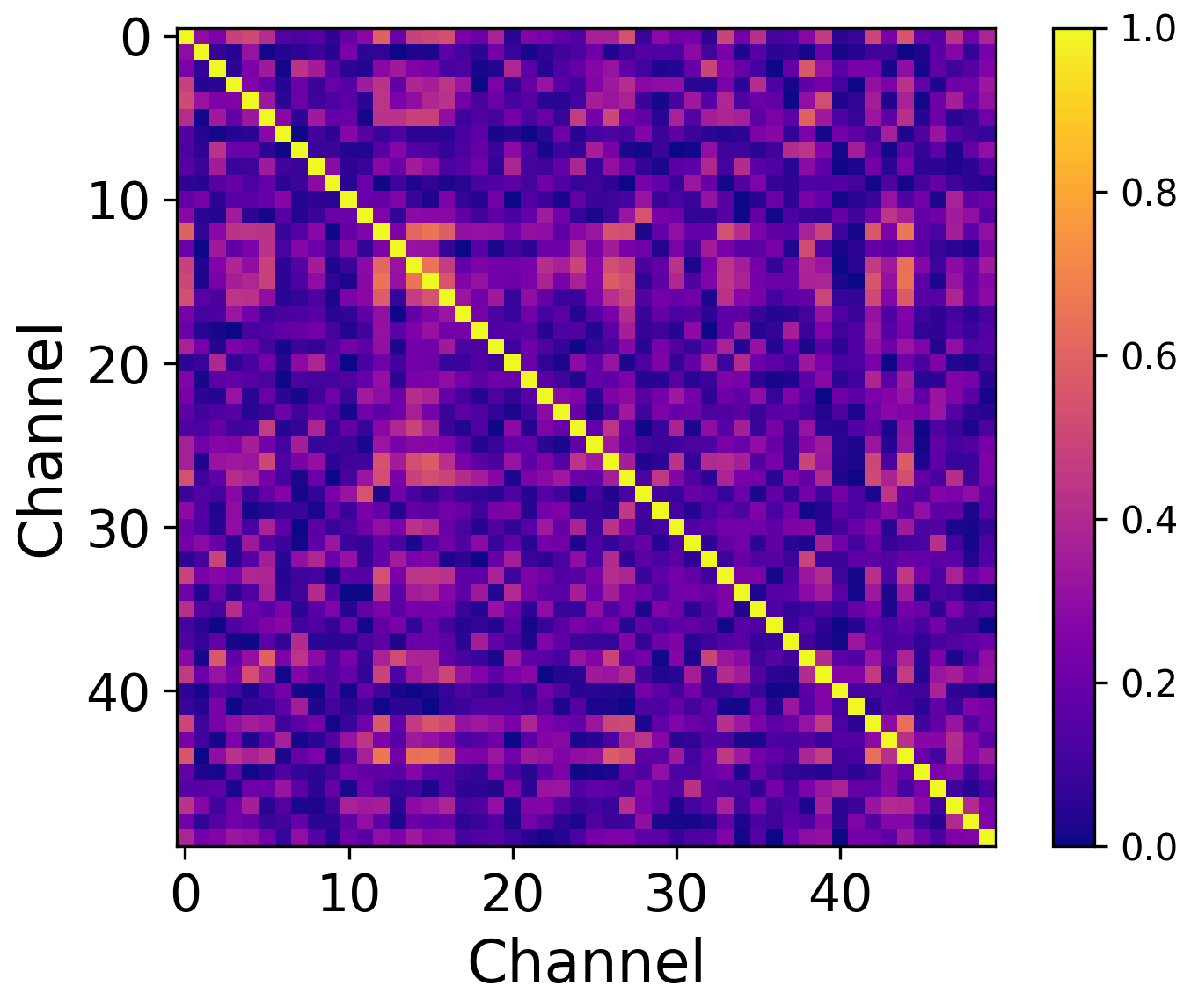}
        \caption{HAC}
    \end{subfigure}
    \begin{subfigure}[b]{0.32\textwidth}
        \includegraphics[width=\linewidth]{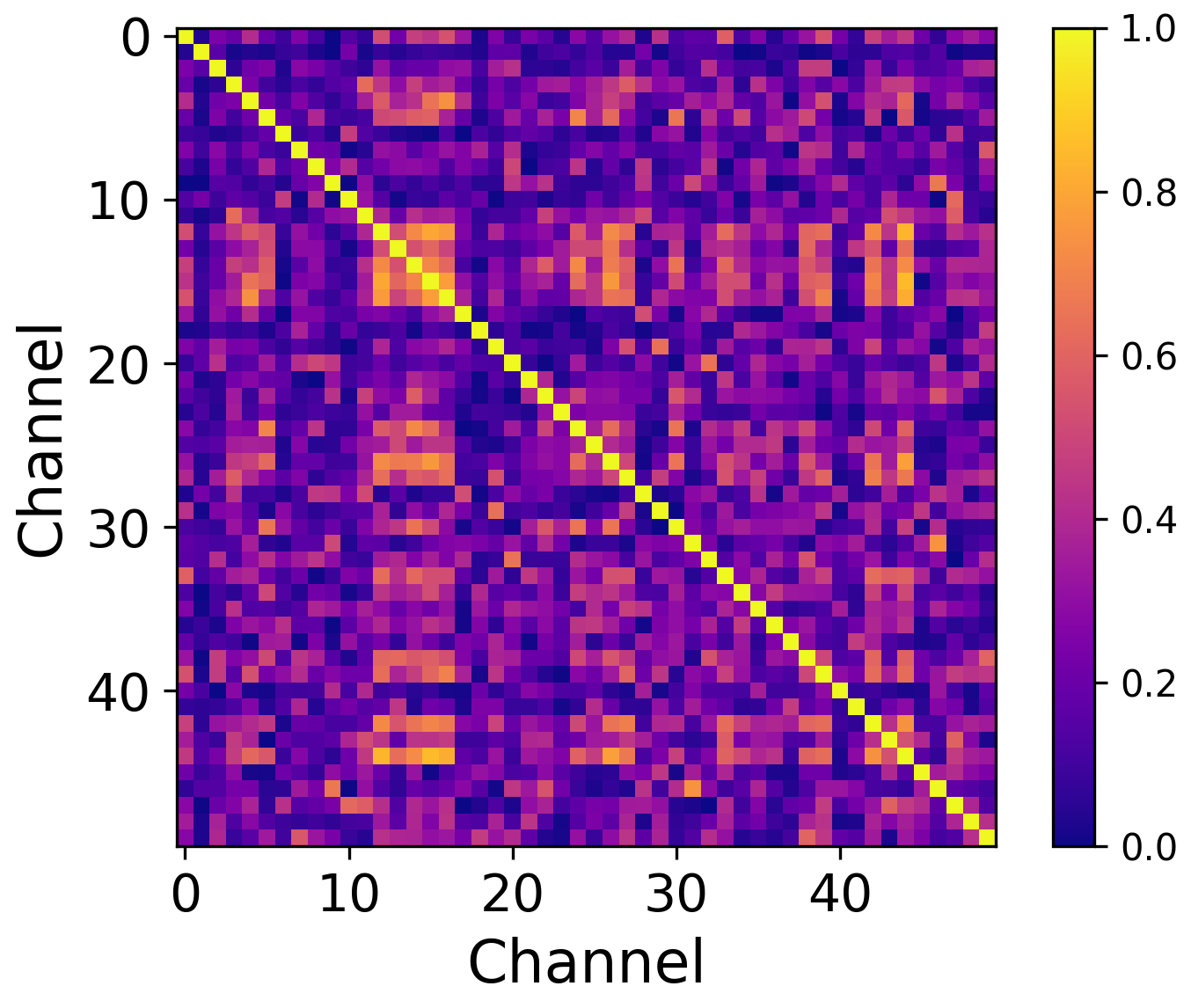}
        \caption{HAC++}
    \end{subfigure}
    \begin{subfigure}[b]{0.32\textwidth}
        \includegraphics[width=\linewidth]{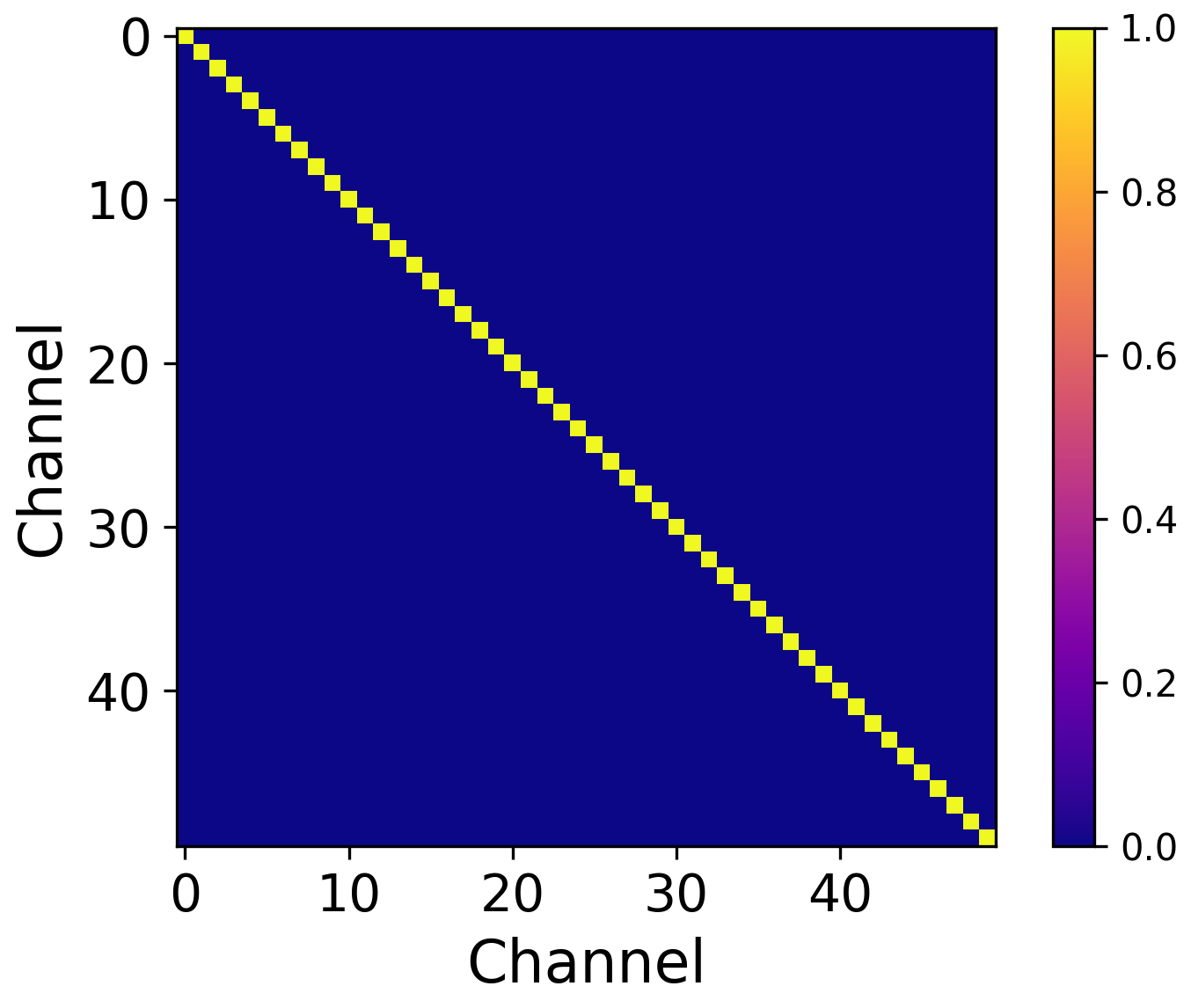}
        \caption{KLT coefficients}
    \end{subfigure}

    \begin{subfigure}[b]{0.32\textwidth}
        \includegraphics[width=\linewidth]{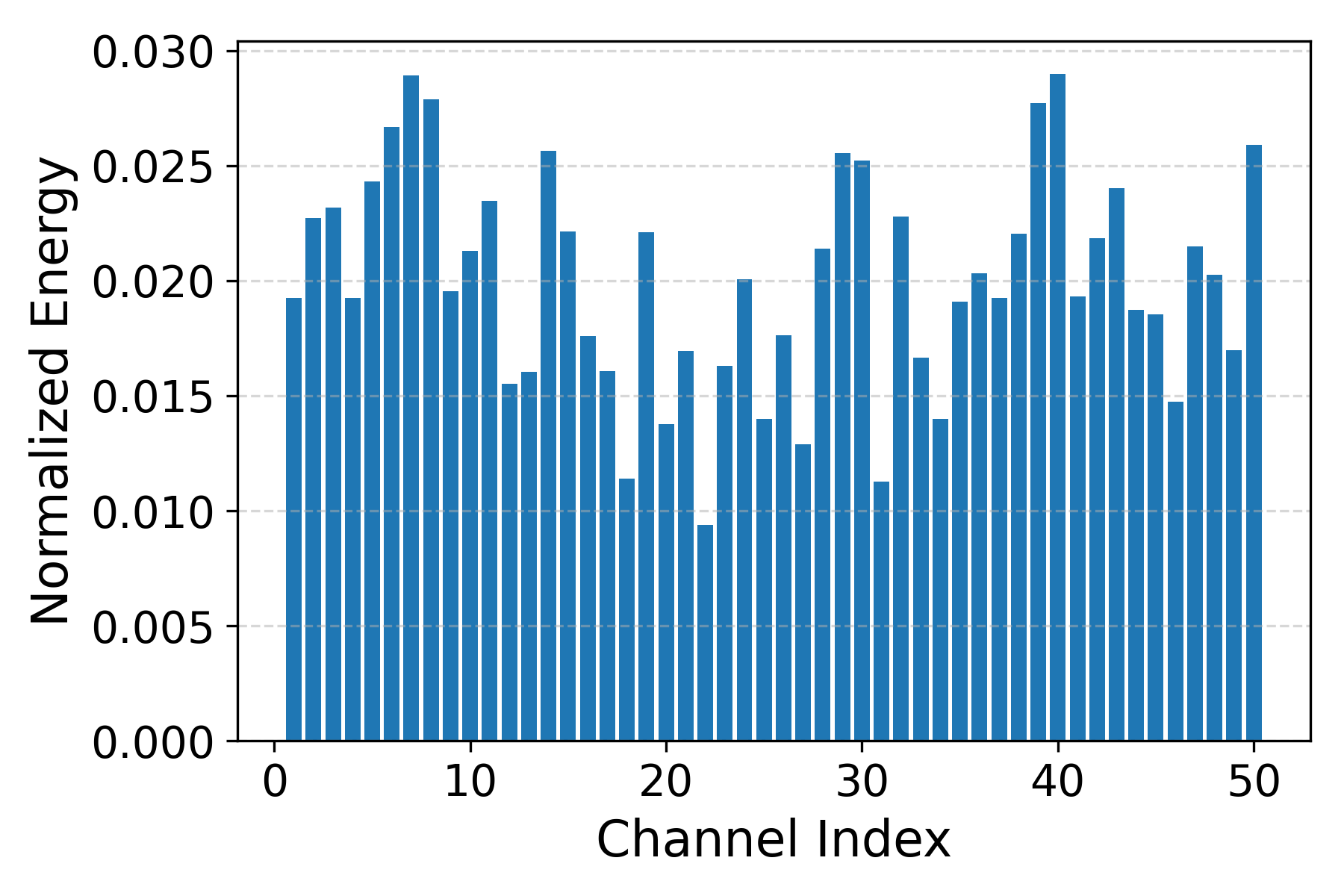}
        \caption{HAC}
    \end{subfigure}
    \begin{subfigure}[b]{0.32\textwidth}
        \includegraphics[width=\linewidth]{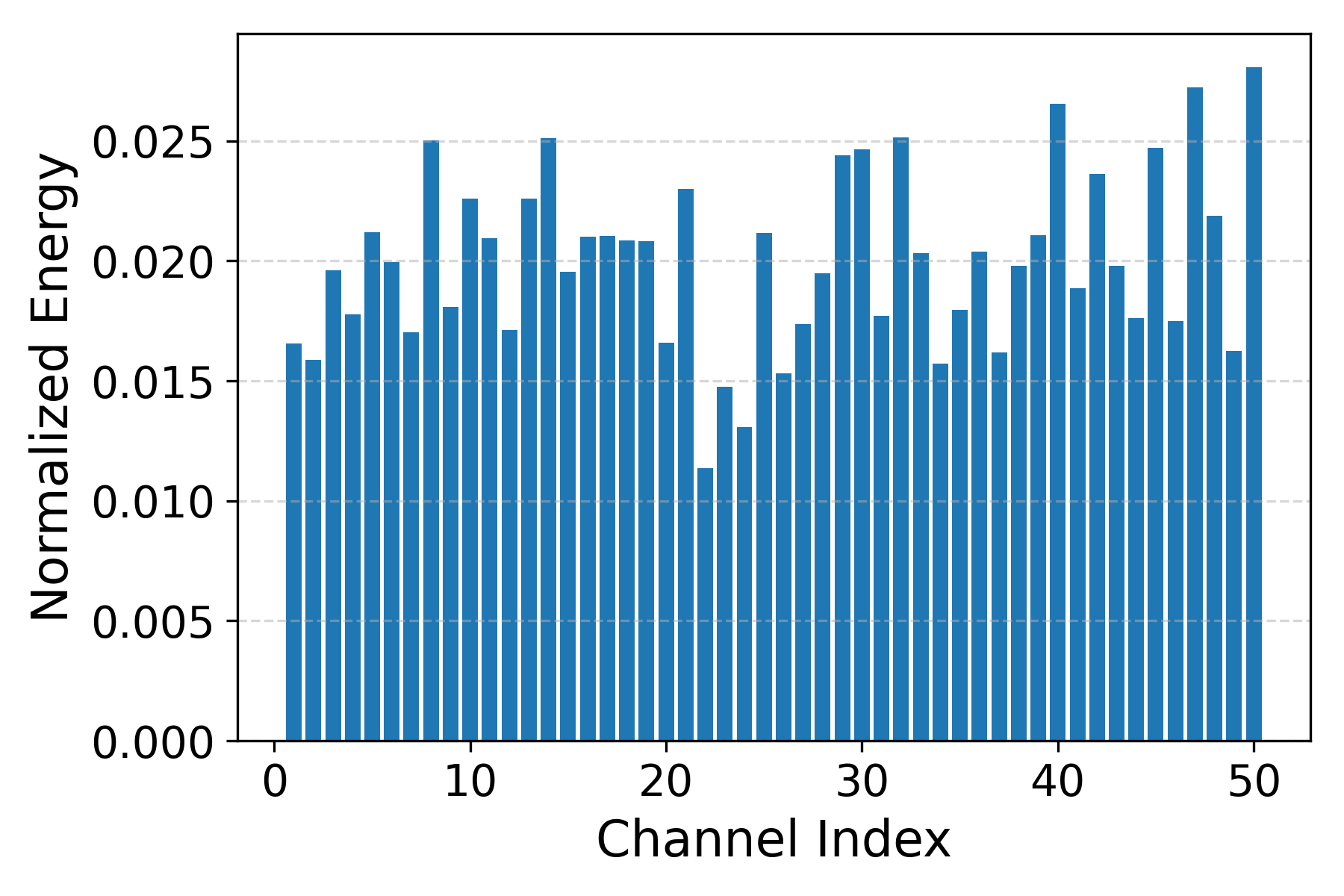}
        \caption{HAC++}
    \end{subfigure}
    \begin{subfigure}[b]{0.32\textwidth}
        \includegraphics[width=\linewidth]{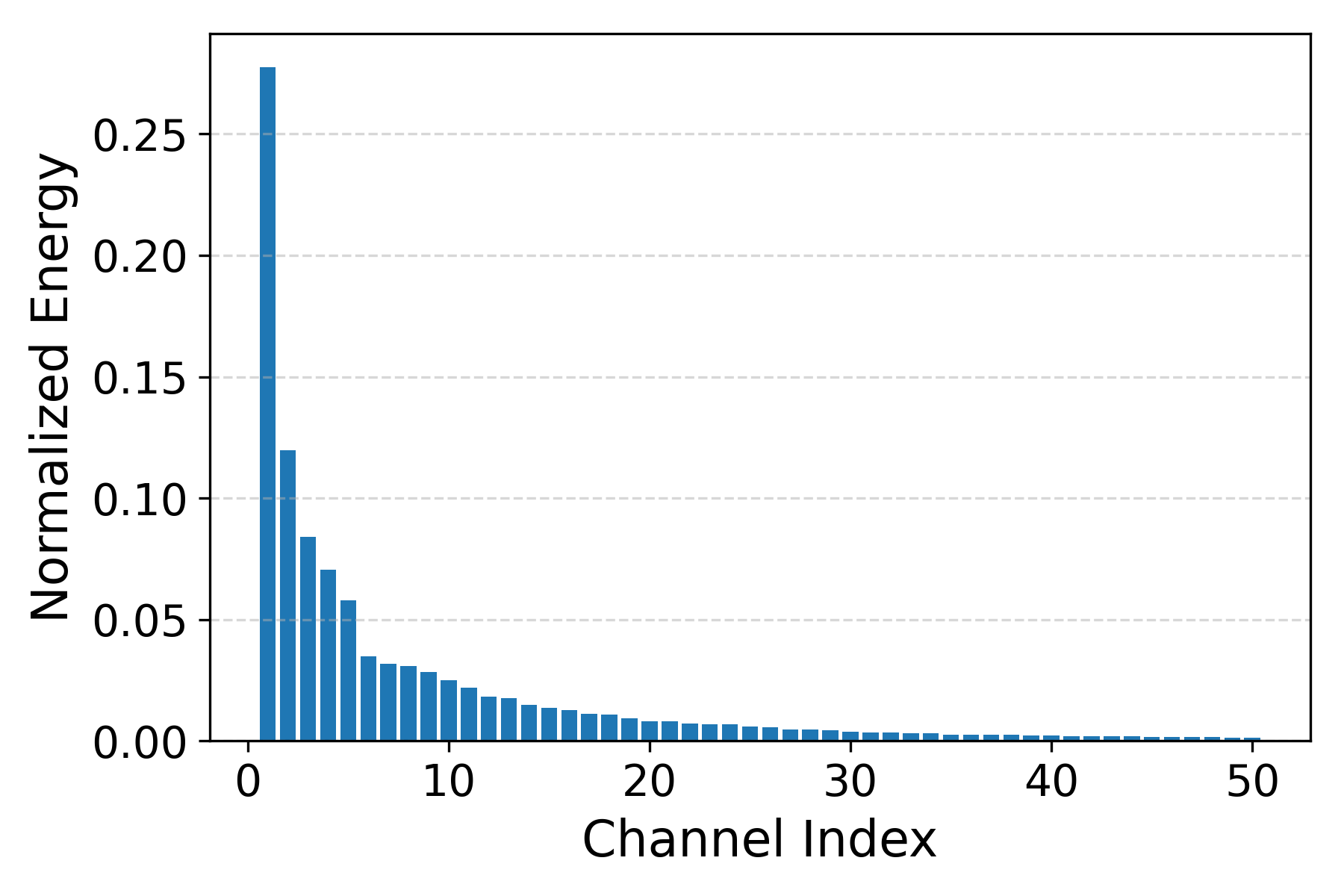}
        \caption{KLT coefficients}
    \end{subfigure}
    \caption{Comparison of inter-channel correlation (visualized as absolute Pearson correlation matrices) and energy compaction in the anchor features of HAC/HAC++ and their KLT coefficients, using the `playroom' scene as an example.}
    \label{fig:correlation_energy}
\end{figure*}

\section{Method}
This section outlines the motivation and challenges for transitioning from anchor-based compression and PTC to our proposed TTC paradigm. We then present our main technical contribution, SHTC.
\subsection{Motivation}
\paragraph{Decorrelation} Let $\mathbf{x}$ denote the anchor-attribute vector to be compressed. Under ideal entropy coding, the minimum achievable rate equals the source entropy $H_q=\mathbb{E}_{\mathbf{x}\sim q(\mathbf{x})}[-\log q(\mathbf{x})]$, but the true distribution $q(\mathbf{x})$ is typically unknown. In practice, one can approximate the unknown distribution $q(\cdot)$ with a parametric model $p(\cdot)$ (often modeled as Gaussian), leading to the cross-entropy rate
\begin{equation}
    R=\mathbb{E}_{\mathbf{x}\sim q(\mathbf{x})}[-\log p(\mathbf{x})]=H_q+D_{KL}(q||p)
\end{equation}
where $D_{KL}(q||p)$ measures the extra bits due to model mismatch.
Anchor attributes exhibit strong statistical dependencies; assuming independence across dimensions/channels therefore increases mismatch and inflates the rate. Existing anchor-based methods mitigate this issue by using correlation-aware entropy models (e.g., ChARM or SARM) to better capture dependencies and reduce $D_{KL}$. 
However, complex entropy models typically incur higher decoding cost (often due to sequential/context-dependent decoding), which is undesirable in 3DGS's ``compress once, decompress many times'' setting where client-side decoding latency is critical. Moreover, fully modeling complex dependencies remains challenging even with sophisticated entropy models, limiting compression gains. A complementary approach is to introduce an analysis transform and decorrelate $\mathbf{x}$ before entropy coding. By reducing redundancy in the representation, we can rely on simpler and faster entropy models without sacrificing rate performance. As illustrated in the first row of Fig.~\ref{fig:correlation_energy}, HAC and HAC++ features show pronounced inter-channel correlations; a more complex entropy model (as in HAC++) models these dependencies but does not remove them from the signal. In contrast, applying a channel-wise KLT substantially suppresses inter-channel correlations, producing near-decorrelated coefficients and paving the way for the use of simpler entropy models. 

\paragraph{Energy compaction} In the original signal domain, the relative importance of channels is unclear, hindering efficient rate allocation. After an analysis transform, energy concentrates in a small subset of coefficients, making their importance explicit and enabling more effective rate allocation. This effect is evident in the second row of Fig.~\ref{fig:correlation_energy}, where KLT compacts most of the energy into the principal coefficients.

\paragraph{PTC vs. TTC: decoupled vs. co-adaptive optimization.} 
Applying an analysis transform before entropy coding can decorrelate the signal and concentrate its energy into fewer coefficients, which improves the R-D efficiency of lossy compression. A paired synthesis transform then reconstructs the anchor attributes for rendering.
A few prior works have also explored analysis-ynthesis transforms~\cite{morgenstern2024compact, lee2025compression}, but they typically follow the PTC paradigm (see Fig.~\ref{fig:Teaser}): applying a fixed, pre-defined transform codec to a frozen 3DGS model in the post-training stage (details in Sec.~\ref{sec:3dgs_post_training_transform}). This two-stage, decoupled design prevents mutual adaptation between the transform and the underlying 3DGS representation, often leading to sub-optimal R-D performance.
In contrast, we advocate the TTC paradigm, where the transforms and entropy model are trained jointly with the 3DGS representation, allowing them to co-adapt and yielding a more compressible representation without sacrificing rendering quality.
\subsection{Challenges and Design Considerations}
TTC is an open-ended paradigm with many candidate transforms (e.g., DCT/DWT, generic MLPs, and graph-based spatial transforms). A natural question then arises: what kinds of transforms are appropriate? To answer this question, we need to understand two domain-specific challenges in 3DGS compression.

The first challenge stems from the irregular, unordered nature of anchor/Gaussian primitives, which lack a natural neighborhood structure. Although neighborhoods can be defined via K-Nearest Neighbors (KNN), neighbor searches in dense point clouds are expensive. Moreover, anchor/Gaussian positions are optimization variables and thus change during training, precluding a one-time KNN construction and requiring repeated graph rebuilding, which substantially increases training cost. This makes graph-based spatial transforms prohibitively expensive to integrate into end-to-end 3DGS training.

The second challenge arises from a fundamental difference between 3DGS compression and neural image/video codecs~\cite{balle2018variational,minnen2018joint,cheng2020learned,he2022elic,li2024neural}. Standard neural codecs employ a single analysis-synthesis transform shared across inputs, and its parameters are not included in the bitstream; this allows the use of parameter-heavy transforms to improve the R--D performance. In contrast, under the TTC paradigm, 3DGS relies on scene-specific transforms whose parameters must be transmitted, imposing a strict parameter budget. Consequently, R-D performance hinges on architectural efficiency and cannot be improved simply by increasing the number of model parameters. Under a tight parameter budget, purely linear transforms often lack sufficient expressive power, while purely MLP-based transforms may be ineffective without suitable inductive bias.

In light of these constraints, we propose SHTC. SHTC circumvents the first challenge by restricting the transform to be channel-wise, i.e., without spatial interactions, making it computation-friendly. To remain effective under a low parameter budget, SHTC adopts a hierarchical architecture, with interpretable components. Details are provided in Sec.~\ref{sec:system}.

\subsection{SHTC: Sparsity-guided Hierarchical Transform Coding}
\label{sec:system}
Transform coding targets decorrelation and energy compaction; among orthonormal linear transforms, KLT is optimal for both. 
Under a tight parameter budget, we also observe that an MLP-based transform fails to outperform the KLT (see Appendix~\ref{sec:transform_choices}). Overall, the KLT offers an excellent cost–performance trade-off, achieving strong R-D performance with minimal compute and memory. Therefore, we use the KLT as the foundation for constructing our transform.
The KLT coefficients exhibit an uneven energy distribution. Discarding low-energy coefficients causes information loss. The artifacts introduced by this truncation become more pronounced at high rates, leading to a quality bottleneck. However, coding all coefficients is rate-inefficient. 
To balance these two extremes, we propose a hierarchical framework integrating KLT with a neural refinement layer.
\begin{itemize}
    \item Layer 1: We use the KLT to construct a base layer for coarse reconstruction. Let $\mathbf{f}\in\mathbb{R}^{N_f}$ denote the anchor feature, with global mean $\mathbf{m}\in\mathbb{R}^{N_f}$ and KLT basis $\mathbf{V}\in\mathbb{R}^{N_f\times N_f}$. The base-layer analysis transform is given by $\boldsymbol{\theta} = g_a(\mathbf{f}) = \mathbf{V}^{\top}(\mathbf{f}-\mathbf{m})$. For the anchor feature, we retain only the top-$M$ principal coefficients $\boldsymbol{\theta}_p=\boldsymbol{\theta}_{1:M}$, which are quantized and entropy-coded. The base-layer synthesis transform is given by  $\hat{\mathbf{f}}_{base}=g_s(\hat{\boldsymbol{\theta}}_p)=\mathbf{V}_{:,1:M}\hat{\boldsymbol{\theta}}_p+\mathbf{m}$.
    \item Layer 2: To compensate for the truncation error of the base layer without incurring significant rate overhead, we introduce a neural refinement layer to transmit the KLT residual. Because most of the signal energy is captured by the leading KLT coefficients, the residual is is typically low-magnitude and approximately sparse/compressible. This observation raises a natural question: can we leverage this sparsity prior to introduce an inductive bias, enabling a very small refinement network to perform effective residual coding with strong R–D performance? 
    We answer this question affirmatively by drawing on classical compressed sensing (CS), which suggests that a high-dimensional sparse (or compressible) signal can often be reconstructed from a small number of linear measurements via sparsity-regularized recovery~\cite{donoho2006compressed}. Motivated by this insight, we design the refinement transforms in a sparsity-aware, CS-inspired manner. Specifically, the refinement-layer analysis transform produces a compact set of learned linear measurements of the residual, while the refinement-layer synthesis transform reconstructs the residual from the decoded measurements using an ISTA-style deep unfolding approach under a sparsity prior~\cite{zhang2018ista,zhang2020deep}. A detailed description of these two transforms is provided next.
\end{itemize}
\subsubsection{Analysis Transform: Learned Linear Measurements of Residuals}
Let $\mathbf{r}$ denote the KLT residual. We encode $\mathbf{r}$ using a learnable linear analysis transform $h_a(\cdot)$, which projects $\mathbf{r}$ into a lower-dimensional vector $\mathbf{y} \in \mathbb{R}^{N_t}$:
\begin{equation}
    \mathbf{y} = h_a(\mathbf{r}) = \mathbf{A}\mathbf{r},
\end{equation}
where $\mathbf{A}$ is a learnable matrix. In the language of compressed sensing, $\mathbf{A}$ implements learned linear measurements of the residual. The measurement vector $\mathbf{y}$ is then quantized to $\hat{\mathbf{y}}$ and entropy coded.
\subsubsection{Synthesis Transform: Residual Reconstruction}
Given the decoded quantized measurements $\hat{\mathbf{y}}$, the synthesis transform $h_s(\cdot)$ reconstructs an estimate $\hat{\mathbf{r}}$ of the residual:
\begin{equation}
    \hat{\mathbf{r}} = h_s(\hat{\mathbf{y}}).
\end{equation}
Rather than designing $h_s(\cdot)$ as a black-box neural network, we adopt an interpretable formulation that casts residual reconstruction as a sparsity-regularized inverse problem. Assuming that $\mathbf{r}$ admits a sparse representation under a learned dictionary $\mathbf{D}$, i.e., $\mathbf{r} \approx \mathbf{D}\boldsymbol{\beta}$ with sparse coefficients $\boldsymbol{\beta}$, the residual reconstruction can be formulated as the following optimization problem:
\begin{equation}
    \tilde{\boldsymbol{\beta}}
    = \arg\min_{\boldsymbol{\beta}}
    \frac{1}{2}\big\|\hat{\mathbf{y}} - \mathbf{A}\mathbf{D}\boldsymbol{\beta}\big\|_2^2
    + \gamma \big\|\boldsymbol{\beta}\big\|_1.
    \label{eq:inv2}
\end{equation}
where the estimation of residual is recovered as $\tilde{\mathbf{r}}=\mathbf{D}\tilde{\boldsymbol{\beta}}$. 

This inverse problem can be addressed using a variety of well-established optimization algorithms. A classical choice is the Iterative Shrinkage-Thresholding Algorithm (ISTA)~\cite{zibulevsky2010l1}. At the $k$-th iteration, ISTA updates the current estimate by performing a gradient descent step on the quadratic data fidelity term, followed by a soft-thresholding operation to handle the non-smooth $\ell_{1}$ regularization term. The update rule is given by:
\begin{equation}
    \boldsymbol{\beta}^{(k+1)}=\mathcal{S}_\tau\big(\boldsymbol{\beta}^{(k)}-\eta\mathbf{D}^T\mathbf{A}^T(\mathbf{A}\mathbf{D}\boldsymbol{\beta}^{(k)}-\hat{\mathbf{y}})\big)
\end{equation}
where $\eta$ denotes the step size, and $\mathcal{S}_\tau(\cdot)$ is the element-wise soft-thresholding function, defined as $\mathcal{S}_\tau(z)=\mathrm{sign}(z)\mathrm{max}(|z|-\tau,0)$. 
Directly applying ISTA, however, can be computationally expensive, as it typically requires many iterations and manual tuning of the hyperparameters $\tau$ and $\eta$. To address this, we adopt a deep unfolding approach~\cite{gregor2010learning,zhang2018ista,zhang2020deep} that maps a large number of ISTA iterations to a small number of learnable network layers, significantly reducing inference time while preserving interpretability and the sparsity prior. Each layer corresponds to one unfolded ISTA iteration, with layer-wise learnable step sizes and thresholds. In particular, $\tau$ and $\eta$ are learned separately for each channel in each layer. This design allows the synthesis transform to embed a learnable sparse structure, enabling efficient residual reconstruction from a compact representation with minimal overhead. 

Concretely, the synthesis transform $h_s(\cdot)$ takes $\hat{\mathbf{y}}$ as input and initializes the sparse coefficient vector $\boldsymbol{\beta}^{(0)}$. Each unfolded ISTA layer then progressively updates the coefficients as
\begin{equation}
    \boldsymbol{\beta}^{(0)}
    \;\rightarrow\;
    \boldsymbol{\beta}^{(1)}
    \;\rightarrow\; \cdots \;\rightarrow\;
    \boldsymbol{\beta}^{(N_s)},
\end{equation}
where $N_s$ is the number of ISTA layers. The final predicted residual is recovered as
\begin{equation}
    \hat{\mathbf{r}} = \mathbf{D}\boldsymbol{\beta}^{(N_s)}.
\end{equation}
Finally, the predicted residual $\hat{\mathbf{r}}$ is combined with the base layer reconstruction to form the final anchor feature for rendering.

\begin{figure}[ht]
  \begin{center}
    \centerline{\includegraphics[width=\columnwidth]{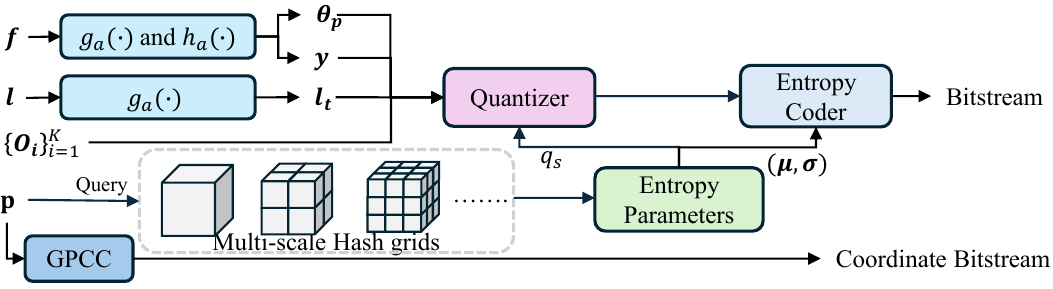}}
    \caption{
      Illustrative system pipeline integrating SHTC into HAC. Anchor coordinates are encoded by MPEG-GPCC, while SHTC latents and other attributes are quantized and entropy-coded using a hash-assisted context model that predicts probability parameters and quantization step sizes conditioned on $\mathbf{p}$.
    }
    \label{fig:system_diagram}
  \end{center}
\end{figure}

\subsection{Implementation details}
We integrate the proposed SHTC into the HAC framework to build our 3DGS compressor. In Fig.~\ref{fig:system_diagram}, we illustrate how the SHTC analysis transform is integrated into the HAC pipeline before quantization and entropy coding. The synthesis transform mirrors the analysis transform and is omitted for brevity. 
Following HAC++~\cite{chen2025hac++}, anchor coordinates $\mathbf{p}$ are compressed using MPEG-GPCC~\cite{graziosi2020overview}$,$ yielding a separate coordinate bitstream.
For the offset vectors $\{\mathbf{O}_i\}_{i=1}^{K}$, we keep the original HAC design. Due to offset masking, the number of valid offsets varies across anchors, making a fixed-dimensional transform impractical; therefore $\{\mathbf{O}_i\}_{i=1}^{K}$ are directly quantized and entropy coded as in HAC.
For the scaling vector $\mathbf{l}$, we use only the base layer of SHTC (without dimensionality reduction) to produce $\mathbf{l}_t \in \mathbb{R}^{6}$, and then apply channel-wise step-size modulation during quantization following Eq.(\ref{eq:quantization_step}) with a dimension-adapted formulation, so that principal coefficients use smaller step sizes than less important ones; the resulting symbols are entropy-coded by HAC.
For the feature vector $\mathbf{f}\in\mathbb{R}^{50}$, the full SHTC produces two latent representations, $\boldsymbol{\theta}_p$ and $\mathbf{y}$, with dimensions $M=15$ and $N_s=15$, respectively; both are quantized and entropy coded by HAC.
As shown in Fig.~\ref{fig:system_diagram}, the probability parameters (e.g., mean and scale) and the quantization step $q_s$ for $\boldsymbol{\theta}_p$, $\mathbf{y}$, $\mathbf{l}_t$ and $\{\mathbf{O}_i\}_{i=1}^{K}$ are predicted from the multi-scale hash grids conditioned on the query position $\mathbf{p}$, and are used by the quantizer and the entropy coder. 

We follow an HAC-style training objective and, in implementation, compute the pixel-wise reconstruction error in YCbCr space rather than RGB. Appendix~\ref{sec:implementation_details} reports the full loss formulation and hyperparameter settings; the YCbCr distortion term is detailed in Appendix~\ref{sec:distortion_term}.

\begin{figure*}[t]
    \centering    
    \begin{subfigure}[b]{0.32\textwidth}
        \includegraphics[width=\linewidth]{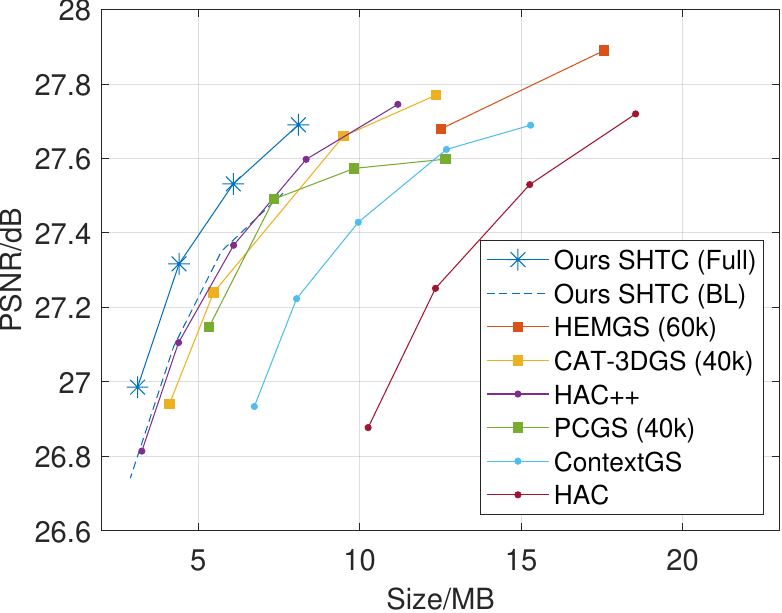}
        \caption{Mip-NeRF360}
    \end{subfigure}
    \begin{subfigure}[b]{0.32\textwidth}
        \includegraphics[width=\linewidth]{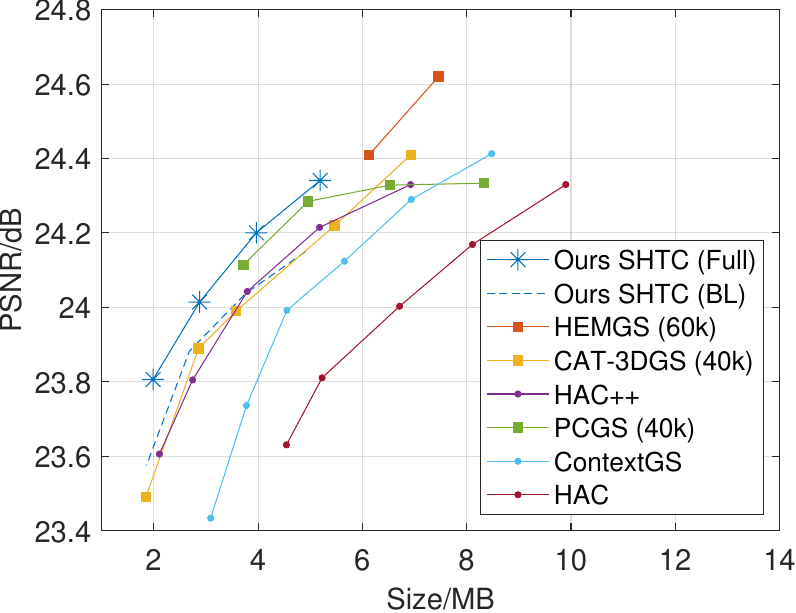}
        \caption{Tank\&Temples}
    \end{subfigure}
    \begin{subfigure}[b]{0.32\textwidth}
        \includegraphics[width=\linewidth]{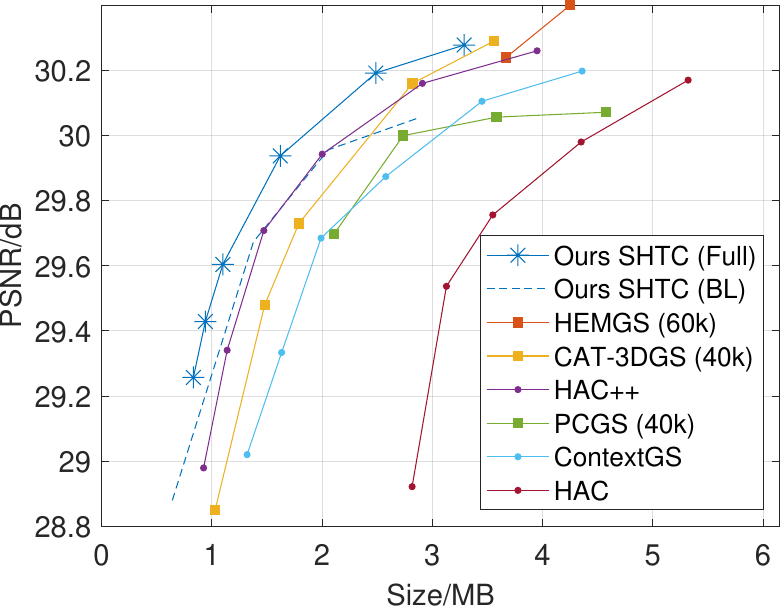}
        \caption{DeepBlending}
    \end{subfigure}
    \caption{Comparison of our method with existing anchor-based 3DGS compression methods.}
    \label{fig:curves_anchor}
\end{figure*}

\section{Experimental Results}
\subsection{Experimental Setup}
We follow the official training protocol of HAC~\cite{chen2024hac} and HAC++~\cite{chen2025hac++}, training each scene for 30{,}000 iterations. All experiments are conducted on a server with 2$\times$ RTX 4090 GPUs. Additional hyperparameter settings are provided in Appendix~\ref{sec:lambda_of_curves}.
\paragraph{Dataset}
We evaluate the R-D performance on five commonly used large-scale real-scene datasets: DeepBlending~\cite{hedman2018deep}, Mip-NeRF360~\cite{barron2022mip}, Tanks\&Temples~\cite{knapitsch2017tanks}, Synthetic-NeRF~\cite{mildenhall2021nerf} and BungeeNeRF~\cite{xiangli2022bungeenerf}. In particular, we assess all nine scenes from the Mip-NeRF360 dataset~\cite{barron2022mip}. Evaluations on additional datasets are provided in the supplementary material. These diverse datasets provide a comprehensive evaluation of the proposed SHTC framework. 
\paragraph{Distortion metrics}
We evaluate rendering distortion using three metrics: PSNR, SSIM~\cite{wang2004image}, and LPIPS~\cite{zhang2018unreasonable}. 
\paragraph{Rate metrics}
Following prior work, we use the storage cost (measured as memory size in MB) as the rate metric.
We adopt the BD-rate~\cite{bjontegaard2001calculation} to quantify rate savings by comparing the R-D curve of a method against a baseline. Specifically, BD-rate reports the average rate difference (in percentage) between two R-D curves over their overlapping distortion range.
A negative BD-rate indicates that the method requires less storage than the baseline at comparable distortion levels, and the magnitude reflects the average percentage of rate reduction.

\subsection{Comparison with Anchor-based Methods}
\subsubsection{Evaluating R-D performance}
\cref{fig:curves_anchor} shows the R-D curves of several anchor-based 3DGS compression methods. Across all three datasets, the curve corresponding to the proposed SHTC, labeled ``Ours SHTC (Full)'' in the legend, lies toward the upper-left region, indicating higher PSNR at a given size or, equivalently, smaller size at a given distortion. 
For HEMGS~\cite{liu2024hemgs}, which is not open-sourced, we plot the R-D points reported in their paper and conduct an approximate comparison. Using only half the training iterations of HEMGS (30k vs. 60k), our method performs on par with or better than HEMGS. 
The R-D curves tested on Synthetic-NeRF and BungeeNeRF are provided in Fig.~\ref{fig:curves_synthetic_bungee_nerf} in Appendix. \cref{tab:our_bd_rate_gains} reports the corresponding BD-rate of our method (HAC + Ours SHTC) relative to HAC++, CAT-3DGS, ContextGS and HAC, quantifying the average memory footprint savings at equal PSNR. These results demonstrate the superiority of our proposed TTC paradigm over anchor-based compression, which relies solely on increasingly complex entropy models.

Note that SHTC is not tailored to HAC and does not depend on it. When integrated into ContextGS, SHTC also yields consistent improvements over the vanilla ContextGS baseline. The R-D curves are shown in Fig.~\ref{fig:context_gs_shtc} in Appendix, and ContextGS+SHTC achieves a BD-rate of $-19.45\%$ relative to vanilla ContextGS.
\begin{table}[htbp]
  \centering
  \caption{The BD-rate of our method relative to four baselines: HAC++, CAT-3DGS, ContextGS and vanilla HAC. Negative values indicate bitrate savings (better) compared with each baseline.}
  \label{tab:our_bd_rate_gains}
  \begin{adjustbox}{width=\linewidth}
  \begin{tabular}{ccccc}
    \toprule
    Dataset & HAC++ &CAT-3DGS (40k) &ContextGS &HAC\\
    \midrule
    Mip-NeRF360    & -20.81\%&-24.54\%&-49.36\% &-64.82\%\\
    Tank\&Temples      & -22.55\%&-22.78\%&-39.17\% &-56.06\%\\
    DeepBlending & -19.58\%&-25.53\%&-42.30\%&-64.56\%\\
    BungeeNeRF      & -10.04\%&-39.90\%&-&-52.05\% \\
    Synthetic-NeRF    & -13.45\%&-&-&-26.33\% \\
    \bottomrule
  \end{tabular}
  \end{adjustbox}
  
\end{table}
\begin{table}[t]
  \caption{Efficiency and rate–distortion comparison on Mip-NeRF360 ($\lambda=0.004$), evaluated on the same RTX 4090 server. HAC is reported as a reference anchor and is not included in the comparison. We compare Ours against ContextGS, CAT-3DGS, and HAC++ in terms of training time (in seconds), rendering FPS, encoding/decoding latency (in seconds), and BD-rate measured relative to the HAC baseline. Best results among the compared methods are highlighted in bold.}
  \label{tab:comp_ours_hac_plus}
  \begin{center}
    \begin{small}
      \begin{sc}
      \begin{adjustbox}{width=\linewidth}
        \begin{tabular}{lccccc}
          \toprule
          \multirow{2}{*}{Methods} & Training  &Rendering & Encoding  & Decoding  & BD-rate  \\
          & Time &FPS& Time &Time &\\
          \midrule
          HAC  & 1949 & 131 & 4.20 & 10.05 & 0 \\
          \hline
          ContextGS& 3927 & 106 & 32.73 & 33.32 &-30.92\% \\
          CAT-3DGS&  5725  & - & 79.82 & 87.95& -53.48\% \\
          HAC++&\textbf{2735}& 128 & 8.48 & 13.86&-56.60\%  \\
          Ours & 3109  & \textbf{146} & \textbf{5.85} &   \textbf{8.84}& \textbf{-64.82\%}      \\
          \bottomrule
        \end{tabular}
        \end{adjustbox}
      \end{sc}
    \end{small}
  \end{center}
  \vskip -0.1in
\end{table}
\subsubsection{Evaluating Transform Overhead}
\paragraph{Computational overhead.}
We use vanilla HAC as the reference baseline and compare our method with state-of-the-art anchor-based codecs, including HAC++, ContextGS, and CAT-3DGS. 
We report BD-rate relative to HAC to quantify R-D gains, and we additionally measure encoding/decoding latency and training time to characterize the cost of these improvements. 
Results are summarized in Table~\ref{tab:comp_ours_hac_plus} and visualized in Fig.~\ref{fig:pareto}. 
Overall, our method achieves improved R--D performance while maintaining low coding latency. 
We attribute this advantage to the introduced transforms, which produce features that better match a lightweight, parallel-friendly entropy model, thereby avoiding expensive and complex entropy modeling.
In terms of training cost, SHTC incurs substantially less overhead than CAT-3DGS and ContextGS, while being moderately slower than HAC++.
Since many 3DGS deployments follow a ``compress once, decompress many times'' workflow, this training overhead is a one-time offline cost for asset producers and typically does not affect client-side user experience.
In contrast, the memory footprint and decoding latency directly impact client-side user experience and are therefore the primary practical constraints.
Consequently, the proposed SHTC method offers a favorable trade-off and lies on (or close to) the empirical Pareto frontier among the compared anchor-based methods as shown in Fig.~\ref{fig:pareto}.

\paragraph{Parameter overhead.} The transforms in SHTC introduce 5,093 additional parameters over the HAC baseline. However, when SHTC is integrated into HAC, the entropy model queries the hash grids and feeds an MLP that predicts probability parameters for low-dimensional transform coefficients instead of high-dimensional anchor features. This change reduces the size of this MLP. As a result, the integration of SHTC into HAC incurs a net overhead of only 1,154 parameters. In contrast, HAC++ increases the number of parameters by 45,400 relative to HAC. These results highlight the parameter efficiency of SHTC, which stems from the sparsity-aware transform design rather than simply adopting a larger black-box MLP.

\subsubsection{Visual Comparison}
Due to space limitations, visual comparisons are provided in Appendix~\ref{sec:add_visual}.

\subsection{Comparison with Other Methods}
In addition to the comparisons with anchor-based compression methods using R-D curves and BD-rate, we also evaluate our method against a broad set of anchor-free and hybrid baselines. Because many of these methods only report one or two operating points on the R-D curve, they do not support a meaningful BD-rate computation; for these methods, we therefore provide a numerical comparison in tabular form in Appendix~\ref{sec:other_method} and Table~\ref{tab:main_results} in Appendix.

In addition, we provide a dedicated comparison with several conceptually representative baselines: the classic 3DGS~\cite{kerbl20233d} and Scaffold-GS~\cite{lu2024scaffold}, the current state-of-the-art (SOTA) anchor-free compression method OMG (NeurIPS 2025)~\cite{lee2025optimized}, the feed-forward compression method FCGS~\cite{chen2025fast}, and four representative methods that apply transform coding in a post-training stage for 3DGS compression: SOG and MesonGS (ECCV 2024)~\cite{morgenstern2024compact,xie2024mesongs}, CodecGS (ICCV 2025)~\cite{lee2025compression}, HybridGS (ICML 2025)~\cite{yang2025hybridgs}. Note that FCGS can also be seen as a method that applies transform coding in a post-training stage after 3DGS is fixed.

As shown in Fig.~\ref{fig:curves_other_methods} in Appendix, our method consistently outperforms the above eight methods. These results clearly demonstrate that jointly optimizing the GS representation, entropy models, and the transforms in an end-to-end manner is substantially more effective than the ad hoc post-training transform paradigm where 3DGS and the transforms cannot be mutually adapted.

\subsection{Ablation Study}
As shown in \cref{fig:curves_anchor}, using only the base layer of SHTC, denoted as `Ours SHTC (BL)', achieves slightly better or comparable R–D performance compared to HAC++ at low and medium bitrates. However, the improvements are modest overall, and SHTC (BL) fails to consistently surpass HAC++ in the high-rate region. This limited gain is mainly due to truncation-induced information loss in the base layer, which inevitably discards some useful information and constrains the achievable performance. In contrast, when we use the full SHTC to compress anchor attributes, which includes an additional neural refinement layer to compensate for truncation-induced information loss, the proposed method yields consistent gains over HAC++ across both low- and high-rate regions. At high bitrates, it also remains clearly superior in terms of R-D performance and does not suffer from the high-rate quality bottleneck observed with the base-layer-only variant.

We conduct a thorough ablation study to isolate the contribution of each component; the main results are reported in Fig.~\ref{fig:ablation_mip_360_all_variants} and Table~\ref{tab:ablation_mip360} in Appendix, with additional details deferred to Appendix~\ref{sec:additional_ablation_study_component}.
In addition to component-wise effects, a natural question is whether residual transform coding is actually necessary for handling truncation. To answer this, we compare (i) our residual transform coding scheme that compensates truncation error against (ii) a direct baseline that transmits all KLT coefficients without truncation. This comparison is provided in Appendix~\ref{sec:additional_ablation_study_no_truncation}.
Finally, since SHTC introduces a new transform design, one may wonder why we adopt this particular design instead of more standard alternatives (e.g., DCT, DWT, or a generic MLP). We therefore include a dedicated analysis of these transform choices in Appendix~\ref{sec:transform_choices}.

Limitations and future work are discussed in Appendix~\ref{sec:limitation_future_work}.
\section{Conclusion}
In this paper, we introduce TTC, a new paradigm for 3DGS compression that jointly optimizes the 3DGS representation, entropy model, and analysis–synthesis transform under a unified R–D objective. Compared with anchor-based compressors that solely rely on increasingly complex entropy models and PTC methods that cannot co-adapt the transform with 3DGS, TTC provides a more effective route to redundancy removal. Within TTC, we propose SHTC, achieving substantial R–D gains and improved inference efficiency with minimal parameter overhead. More broadly, our parameter-efficient design may benefit low-complexity neural image and video codec design.

\section*{Impact Statement}
Our method serves purely as a compression tool, which operates solely on existing data to reduce its storage or transmission cost. It neither generates new data nor edits the original content, thereby minimizing risks of misuse, such as data fabrication or manipulation. Given its functional scope, we do not anticipate any significant negative societal impacts associated with the deployment of our method.

On the positive side, by significantly reducing the file size of 3DGS representations and lowering decoding latency, our method facilitates faster downloading and decoding. This improves user experience in bandwidth-constrained or real-time applications such as virtual reality, immersive gaming, architectural visualization, and cultural heritage preservation. 

\nocite{langley00}

\bibliography{example_paper}

@String(TOG= {ACM Trans. Graph.})

@String(ICASSP=	{ICASSP})

@String(ICIP = {IEEE Int. Conf. Image Process.})

@String(TOG   = {ACM TOG})

@String(ICIP  = {ICIP})

@article{balle2018variational,
  title={Variational image compression with a scale hyperprior},
  author={Ball{\'e}, Johannes and Minnen, David and Singh, Saurabh and Hwang, Sung Jin and Johnston, Nick},
  journal={arXiv preprint arXiv:1802.01436},
  year={2018}
}

@article{minnen2018joint,
  title={Joint autoregressive and hierarchical priors for learned image compression},
  author={Minnen, David and Ball{\'e}, Johannes and Toderici, George D},
  journal={Advances in neural information processing systems},
  volume={31},
  year={2018}
}

@inproceedings{cheng2020learned,
  title={Learned image compression with discretized gaussian mixture likelihoods and attention modules},
  author={Cheng, Zhengxue and Sun, Heming and Takeuchi, Masaru and Katto, Jiro},
  booktitle={Proceedings of the IEEE/CVF conference on computer vision and pattern recognition},
  pages={7939--7948},
  year={2020}
}

@inproceedings{he2022elic,
  title={Elic: Efficient learned image compression with unevenly grouped space-channel contextual adaptive coding},
  author={He, Dailan and Yang, Ziming and Peng, Weikun and Ma, Rui and Qin, Hongwei and Wang, Yan},
  booktitle={Proceedings of the IEEE/CVF Conference on Computer Vision and Pattern Recognition},
  pages={5718--5727},
  year={2022}
}

@article{mildenhall2021nerf,
  title={Nerf: Representing scenes as neural radiance fields for view synthesis},
  author={Mildenhall, Ben and Srinivasan, Pratul P and Tancik, Matthew and Barron, Jonathan T and Ramamoorthi, Ravi and Ng, Ren},
  journal={Communications of the ACM},
  volume={65},
  number={1},
  pages={99--106},
  year={2021},
  publisher={ACM New York, NY, USA}
}

@article{kerbl20233d,
  title={3d gaussian splatting for real-time radiance field rendering.},
  author={Kerbl, Bernhard and Kopanas, Georgios and Leimk{\"u}hler, Thomas and Drettakis, George},
  journal={ACM Trans. Graph.},
  volume={42},
  number={4},
  pages={139--1},
  year={2023}
}

@inproceedings{lu2024scaffold,
  title={Scaffold-gs: Structured 3d gaussians for view-adaptive rendering},
  author={Lu, Tao and Yu, Mulin and Xu, Linning and Xiangli, Yuanbo and Wang, Limin and Lin, Dahua and Dai, Bo},
  booktitle={Proceedings of the IEEE/CVF Conference on Computer Vision and Pattern Recognition},
  pages={20654--20664},
  year={2024}
}

@inproceedings{morgenstern2024compact,
  title={Compact 3d scene representation via self-organizing gaussian grids},
  author={Morgenstern, Wieland and Barthel, Florian and Hilsmann, Anna and Eisert, Peter},
  booktitle={European Conference on Computer Vision},
  pages={18--34},
  year={2024},
  organization={Springer}
}

@article{lee2025compression,
  title={Compression of 3D Gaussian Splatting with Optimized Feature Planes and Standard Video Codecs},
  author={Lee, Soonbin and Shu, Fangwen and Sanchez, Yago and Schierl, Thomas and Hellge, Cornelius},
  journal={arXiv preprint arXiv:2501.03399},
  year={2025}
}

@article{liu2024hemgs,
  title={HEMGS: A Hybrid Entropy Model for 3D Gaussian Splatting Data Compression},
  author={Liu, Lei and Chen, Zhenghao and Xu, Dong},
  journal={arXiv preprint arXiv:2411.18473},
  year={2024}
}

@inproceedings{
zhan2025catdgs,
title={{CAT}-3{DGS}: A Context-Adaptive Triplane Approach to Rate-Distortion-Optimized 3{DGS} Compression},
author={Yu-Ting Zhan and Cheng-Yuan Ho and Hebi Yang and Yi-Hsin Chen and Jui Chiu Chiang and Yu-Lun Liu and Wen-Hsiao Peng},
booktitle={The Thirteenth International Conference on Learning Representations},
year={2025},
url={https://openreview.net/forum?id=m3KuuE2ozw}
}

@inproceedings{chen2024hac,
  title={Hac: Hash-grid assisted context for 3d gaussian splatting compression},
  author={Chen, Yihang and Wu, Qianyi and Lin, Weiyao and Harandi, Mehrtash and Cai, Jianfei},
  booktitle={European Conference on Computer Vision},
  pages={422--438},
  year={2024},
  organization={Springer}
}

@inproceedings{liu2024compgs,
  title={Compgs: Efficient 3d scene representation via compressed gaussian splatting},
  author={Liu, Xiangrui and Wu, Xinju and Zhang, Pingping and Wang, Shiqi and Li, Zhu and Kwong, Sam},
  booktitle={Proceedings of the 32nd ACM International Conference on Multimedia},
  pages={2936--2944},
  year={2024}
}

@inproceedings{
wang2024contextgs,
title={Context{GS} : Compact 3D Gaussian Splatting with Anchor Level Context Model},
author={Yufei Wang and Zhihao Li and Lanqing Guo and Wenhan Yang and Alex Kot and Bihan Wen},
booktitle={The Thirty-eighth Annual Conference on Neural Information Processing Systems},
year={2024},
url={https://openreview.net/forum?id=W2qGSMl2Uu}
}

@inproceedings{girish2024eagles,
  title={Eagles: Efficient accelerated 3d gaussians with lightweight encodings},
  author={Girish, Sharath and Gupta, Kamal and Shrivastava, Abhinav},
  booktitle={European Conference on Computer Vision},
  pages={54--71},
  year={2024},
  organization={Springer}
}

@inproceedings{xie2024mesongs,
  title={Mesongs: Post-training compression of 3d gaussians via efficient attribute transformation},
  author={Xie, Shuzhao and Zhang, Weixiang and Tang, Chen and Bai, Yunpeng and Lu, Rongwei and Ge, Shijia and Wang, Zhi},
  booktitle={European Conference on Computer Vision},
  pages={434--452},
  year={2024},
  organization={Springer}
}

@article{papantonakis2024reducing,
  title={Reducing the Memory Footprint of 3D Gaussian Splatting},
  author={Papantonakis, Panagiotis and Kopanas, Georgios and Kerbl, Bernhard and Lanvin, Alexandre and Drettakis, George},
  journal={Proceedings of the ACM on Computer Graphics and Interactive Techniques},
  volume={7},
  number={1},
  pages={1--17},
  year={2024},
  publisher={ACM New York, NY, USA}
}

@inproceedings{wang2024end,
  title={End-to-end rate-distortion optimized 3d gaussian representation},
  author={Wang, Henan and Zhu, Hanxin and He, Tianyu and Feng, Runsen and Deng, Jiajun and Bian, Jiang and Chen, Zhibo},
  booktitle={European Conference on Computer Vision},
  pages={76--92},
  year={2024},
  organization={Springer}
}

@article{fan2024lightgaussian,
  title={Lightgaussian: Unbounded 3d gaussian compression with 15x reduction and 200+ fps},
  author={Fan, Zhiwen and Wang, Kevin and Wen, Kairun and Zhu, Zehao and Xu, Dejia and Wang, Zhangyang and others},
  journal={Advances in neural information processing systems},
  volume={37},
  pages={140138--140158},
  year={2024}
}

@inproceedings{lee2024compact,
  title={Compact 3d gaussian representation for radiance field},
  author={Lee, Joo Chan and Rho, Daniel and Sun, Xiangyu and Ko, Jong Hwan and Park, Eunbyung},
  booktitle={Proceedings of the IEEE/CVF Conference on Computer Vision and Pattern Recognition},
  pages={21719--21728},
  year={2024}
}

@inproceedings{niedermayr2024compressed,
  title={Compressed 3d gaussian splatting for accelerated novel view synthesis},
  author={Niedermayr, Simon and Stumpfegger, Josef and Westermann, R{\"u}diger},
  booktitle={Proceedings of the IEEE/CVF Conference on Computer Vision and Pattern Recognition},
  pages={10349--10358},
  year={2024}
}

@inproceedings{navaneet2024compgs,
  title={Compgs: Smaller and faster gaussian splatting with vector quantization},
  author={Navaneet, KL and Pourahmadi Meibodi, Kossar and Abbasi Koohpayegani, Soroush and Pirsiavash, Hamed},
  booktitle={European Conference on Computer Vision},
  pages={330--349},
  year={2024},
  organization={Springer}
}

@article{niemeyer2024radsplat,
  title={Radsplat: Radiance field-informed gaussian splatting for robust real-time rendering with 900+ fps},
  author={Niemeyer, Michael and Manhardt, Fabian and Rakotosaona, Marie-Julie and Oechsle, Michael and Duckworth, Daniel and Gosula, Rama and Tateno, Keisuke and Bates, John and Kaeser, Dominik and Tombari, Federico},
  journal={arXiv preprint arXiv:2403.13806},
  year={2024}
}

@article{ali2024elmgs,
  title={ELMGS: Enhancing memory and computation scaLability through coMpression for 3D Gaussian Splatting},
  author={Ali, Muhammad Salman and Bae, Sung-Ho and Tartaglione, Enzo},
  journal={arXiv preprint arXiv:2410.23213},
  year={2024}
}

@article{ali2024trimming,
  title={Trimming the fat: Efficient compression of 3d gaussian splats through pruning},
  author={Ali, Muhammad Salman and Qamar, Maryam and Bae, Sung-Ho and Tartaglione, Enzo},
  journal={arXiv preprint arXiv:2406.18214},
  year={2024}
}

@ARTICLE{chen2025hac++,
  author={Chen, Yihang and Wu, Qianyi and Lin, Weiyao and Harandi, Mehrtash and Cai, Jianfei},
  journal={IEEE Transactions on Pattern Analysis and Machine Intelligence}, 
  title={HAC++: Towards 100X Compression of 3D Gaussian Splatting}, 
  year={2025},
  volume={},
  number={},
  pages={1-17},
  doi={10.1109/TPAMI.2025.3594066}}

@inproceedings{
chen2025fast,
title={Fast Feedforward 3D Gaussian Splatting Compression},
author={Yihang Chen and Qianyi Wu and Mengyao Li and Weiyao Lin and Mehrtash Harandi and Jianfei Cai},
booktitle={The Thirteenth International Conference on Learning Representations},
year={2025},
url={https://openreview.net/forum?id=DCandSZ2F1}
}

@article{knapitsch2017tanks,
  title={Tanks and temples: Benchmarking large-scale scene reconstruction},
  author={Knapitsch, Arno and Park, Jaesik and Zhou, Qian-Yi and Koltun, Vladlen},
  journal={ACM Transactions on Graphics (ToG)},
  volume={36},
  number={4},
  pages={1--13},
  year={2017},
  publisher={ACM New York, NY, USA}
}

@article{hedman2018deep,
  title={Deep blending for free-viewpoint image-based rendering},
  author={Hedman, Peter and Philip, Julien and Price, True and Frahm, Jan-Michael and Drettakis, George and Brostow, Gabriel},
  journal={ACM Transactions on Graphics (ToG)},
  volume={37},
  number={6},
  pages={1--15},
  year={2018},
  publisher={ACM New York, NY, USA}
}

@inproceedings{barron2022mip,
  title={Mip-nerf 360: Unbounded anti-aliased neural radiance fields},
  author={Barron, Jonathan T and Mildenhall, Ben and Verbin, Dor and Srinivasan, Pratul P and Hedman, Peter},
  booktitle={Proceedings of the IEEE/CVF conference on computer vision and pattern recognition},
  pages={5470--5479},
  year={2022}
}

@article{graziosi2020overview,
  title={An overview of ongoing point cloud compression standardization activities: Video-based (V-PCC) and geometry-based (G-PCC)},
  author={Graziosi, Danillo and Nakagami, Ohji and Kuma, Shinroku and Zaghetto, Alexandre and Suzuki, Teruhiko and Tabatabai, Ali},
  journal={APSIPA Transactions on Signal and Information Processing},
  volume={9},
  pages={e13},
  year={2020},
  publisher={Cambridge University Press}
}

@inproceedings{chan2022efficient,
  title={Efficient geometry-aware 3d generative adversarial networks},
  author={Chan, Eric R and Lin, Connor Z and Chan, Matthew A and Nagano, Koki and Pan, Boxiao and De Mello, Shalini and Gallo, Orazio and Guibas, Leonidas J and Tremblay, Jonathan and Khamis, Sameh and others},
  booktitle={Proceedings of the IEEE/CVF conference on computer vision and pattern recognition},
  pages={16123--16133},
  year={2022}
}

@inproceedings{minnen2020channel,
  title={Channel-wise autoregressive entropy models for learned image compression},
  author={Minnen, David and Singh, Saurabh},
  booktitle={2020 IEEE International Conference on Image Processing (ICIP)},
  pages={3339--3343},
  year={2020},
  organization={IEEE}
}

@article{bjontegaard2001calculation,
  title={Calculation of average PSNR differences between RD-curves},
  author={Bjontegaard, Gisle},
  journal={ITU SG16 Doc. VCEG-M33},
  year={2001}
}

@article{wang2004image,
  title={Image quality assessment: from error visibility to structural similarity},
  author={Wang, Zhou and Bovik, Alan C and Sheikh, Hamid R and Simoncelli, Eero P},
  journal={IEEE transactions on image processing},
  volume={13},
  number={4},
  pages={600--612},
  year={2004},
  publisher={IEEE}
}

@inproceedings{zhang2018unreasonable,
  title={The unreasonable effectiveness of deep features as a perceptual metric},
  author={Zhang, Richard and Isola, Phillip and Efros, Alexei A and Shechtman, Eli and Wang, Oliver},
  booktitle={Proceedings of the IEEE conference on computer vision and pattern recognition},
  pages={586--595},
  year={2018}
}

@inproceedings{yang2024benchmark,
  title={A benchmark for gaussian splatting compression and quality assessment study},
  author={Yang, Qi and Yang, Kaifa and Xing, Yuke and Xu, Yiling and Li, Zhu},
  booktitle={Proceedings of the 6th ACM International Conference on Multimedia in Asia},
  pages={1--8},
  year={2024}
}

@inproceedings{huang2025hierarchical,
  title={A hierarchical compression technique for 3d gaussian splatting compression},
  author={Huang, He and Huang, Wenjie and Yang, Qi and Xu, Yiling and Li, Zhu},
  booktitle={ICASSP 2025-2025 IEEE International Conference on Acoustics, Speech and Signal Processing (ICASSP)},
  pages={1--5},
  year={2025},
  organization={IEEE}
}

@inproceedings{sandryhaila2013discrete,
  title={Discrete signal processing on graphs: Graph fourier transform},
  author={Sandryhaila, Aliaksei and Moura, Jos{\'e} MF},
  booktitle={2013 IEEE International Conference on Acoustics, Speech and Signal Processing},
  pages={6167--6170},
  year={2013},
  organization={IEEE}
}

@article{de2016compression,
  title={Compression of 3D point clouds using a region-adaptive hierarchical transform},
  author={De Queiroz, Ricardo L and Chou, Philip A},
  journal={IEEE Transactions on Image Processing},
  volume={25},
  number={8},
  pages={3947--3956},
  year={2016},
  publisher={IEEE}
}

@inproceedings{zhang2020deep,
  title={Deep unfolding network for image super-resolution},
  author={Zhang, Kai and Gool, Luc Van and Timofte, Radu},
  booktitle={Proceedings of the IEEE/CVF conference on computer vision and pattern recognition},
  pages={3217--3226},
  year={2020}
}

@inproceedings{zhang2018ista,
  title={ISTA-Net: Interpretable optimization-inspired deep network for image compressive sensing},
  author={Zhang, Jian and Ghanem, Bernard},
  booktitle={Proceedings of the IEEE conference on computer vision and pattern recognition},
  pages={1828--1837},
  year={2018}
}

@article{zibulevsky2010l1,
  title={L1-L2 optimization in signal and image processing},
  author={Zibulevsky, Michael and Elad, Michael},
  journal={IEEE Signal Processing Magazine},
  volume={27},
  number={3},
  pages={76--88},
  year={2010},
  publisher={IEEE}
}

@inproceedings{li2024neural,
  title={Neural video compression with feature modulation},
  author={Li, Jiahao and Li, Bin and Lu, Yan},
  booktitle={Proceedings of the IEEE/CVF Conference on Computer Vision and Pattern Recognition},
  pages={26099--26108},
  year={2024}
}

@article{donoho2006compressed,
  title={Compressed sensing},
  author={Donoho, David L},
  journal={IEEE Transactions on information theory},
  volume={52},
  number={4},
  pages={1289--1306},
  year={2006},
  publisher={IEEE}
}

@inproceedings{hanson2025pup,
  title={Pup 3d-gs: Principled uncertainty pruning for 3d gaussian splatting},
  author={Hanson, Alex and Tu, Allen and Singla, Vasu and Jayawardhana, Mayuka and Zwicker, Matthias and Goldstein, Tom},
  booktitle={Proceedings of the Computer Vision and Pattern Recognition Conference},
  pages={5949--5958},
  year={2025}
}

@article{wang2025adaptive,
  title={Adaptive Voxelization for Transform coding of 3D Gaussian splatting data},
  author={Wang, Chenjunjie and Sridhara, Shashank N and Pavez, Eduardo and Ortega, Antonio and Chang, Cheng},
  journal={arXiv preprint arXiv:2506.00271},
  year={2025}
}

@inproceedings{
shin2025localityaware,
title={Locality-aware Gaussian Compression for Fast and High-quality Rendering},
author={Seungjoo Shin and Jaesik Park and Sunghyun Cho},
booktitle={The Thirteenth International Conference on Learning Representations},
year={2025},
url={https://openreview.net/forum?id=dHYwfV2KeP}
}

@inproceedings{xie2025sizegs,
  title={SizeGS: Size-aware Compression of 3D Gaussian Splatting via Mixed Integer Programming},
  author={Xie, Shuzhao and Liu, Jiahang and Zhang, Weixiang and Ge, Shijia and Pan, Sicheng and Tang, Chen and Bai, Yunpeng and Zhang, Cong and Fan, Xiaoyi and Wang, Zhi},
  booktitle={Proceedings of the 33rd ACM International Conference on Multimedia},
  pages={8214--8223},
  year={2025}
}

@inproceedings{tian2025flexgaussian,
  title={Flexgaussian: Flexible and cost-effective training-free compression for 3d gaussian splatting},
  author={Tian, Boyuan and Gao, Qizhe and Xianyu, Siran and Cui, Xiaotong and Zhang, Minjia},
  booktitle={Proceedings of the 33rd ACM International Conference on Multimedia},
  pages={7287--7296},
  year={2025}
}

@article{tang2025neuralgs,
  title={NeuralGS: Bridging Neural Fields and 3D Gaussian Splatting for Compact 3D Representations},
  author={Tang, Zhenyu and Feng, Chaoran and Cheng, Xinhua and Yu, Wangbo and Zhang, Junwu and Liu, Yuan and Long, Xiaoxiao and Wang, Wenping and Yuan, Li},
  journal={arXiv preprint arXiv:2503.23162},
  year={2025}
}

@article{ma2025enhancing,
  title={Enhancing 3D Gaussian Splatting Compression via Spatial Condition-based Prediction},
  author={Ma, Jingui and Hu, Yang and Tang, Luyang and Yang, Jiayu and Zhai, Yongqi and Wang, Ronggang},
  journal={arXiv preprint arXiv:2503.23337},
  year={2025}
}

@inproceedings{tang2025feature,
  title={Feature Prediction for 3D Gaussian Splatting Compression},
  author={Tang, Luyang and Zhai, Yongqi and Yang, Jiayu and Yang, Chunhui and Wang, Ronggang},
  booktitle={2025 Data Compression Conference (DCC)},
  pages={73--82},
  year={2025},
  organization={IEEE}
}

@inproceedings{
yang2025hybridgs,
title={Hybrid{GS}: High-Efficiency Gaussian Splatting Data Compression using Dual-Channel Sparse Representation and Point Cloud Encoder},
author={Qi Yang and Le Yang and Geert Van der Auwera and Zhu Li},
booktitle={Forty-second International Conference on Machine Learning},
year={2025},
url={https://openreview.net/forum?id=6mQv4fnsj0}
}

@inproceedings{zhang2025sogs,
  title={SOGS: Second-Order Anchor for Advanced 3D Gaussian Splatting},
  author={Zhang, Jiahui and Zhan, Fangneng and Shao, Ling and Lu, Shijian},
  booktitle={Proceedings of the Computer Vision and Pattern Recognition Conference},
  pages={11167--11176},
  year={2025}
}

@article{lee2025optimized,
  title={Optimized Minimal 3D Gaussian Splatting},
  author={Lee, Joo Chan and Ko, Jong Hwan and Park, Eunbyung},
  journal={arXiv preprint arXiv:2503.16924},
  year={2025}
}

@article{chen2025pcgs,
  title={Pcgs: Progressive compression of 3d gaussian splatting},
  author={Chen, Yihang and Li, Mengyao and Wu, Qianyi and Lin, Weiyao and Harandi, Mehrtash and Cai, Jianfei},
  journal={arXiv preprint arXiv:2503.08511},
  year={2025}
}

@inproceedings{gregor2010learning,
  title={Learning fast approximations of sparse coding},
  author={Gregor, Karol and LeCun, Yann},
  booktitle={Proceedings of the 27th international conference on international conference on machine learning},
  pages={399--406},
  year={2010}
}

@inproceedings{xiangli2022bungeenerf,
  title={Bungeenerf: Progressive neural radiance field for extreme multi-scale scene rendering},
  author={Xiangli, Yuanbo and Xu, Linning and Pan, Xingang and Zhao, Nanxuan and Rao, Anyi and Theobalt, Christian and Dai, Bo and Lin, Dahua},
  booktitle={European conference on computer vision},
  pages={106--122},
  year={2022},
  organization={Springer}
}

@article{rissanen1978modeling,
  title={Modeling by shortest data description},
  author={Rissanen, Jorma},
  journal={Automatica},
  volume={14},
  number={5},
  pages={465--471},
  year={1978},
  publisher={Elsevier}
}

@article{elsken2019neural,
  title={Neural architecture search: A survey},
  author={Elsken, Thomas and Metzen, Jan Hendrik and Hutter, Frank},
  journal={Journal of Machine Learning Research},
  volume={20},
  number={55},
  pages={1--21},
  year={2019}
}
\bibliographystyle{icml2026}

\newpage
\appendix
\onecolumn

\begin{figure}[t]
    \centering    
    \begin{subfigure}[b]{0.45\textwidth}
        \includegraphics[width=0.9\linewidth]{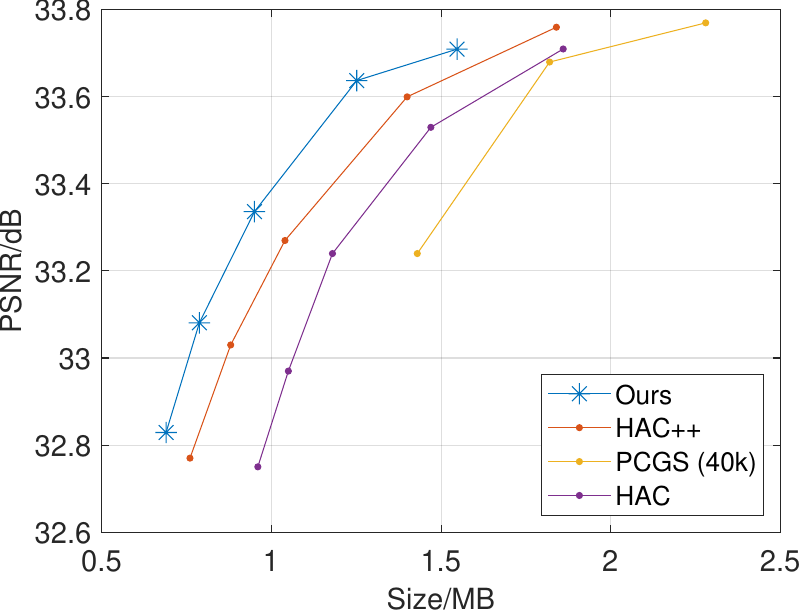}
        \caption{Synthetic-NeRF}
    \end{subfigure}
    \begin{subfigure}[b]{0.45\textwidth}
        \includegraphics[width=0.9\linewidth]{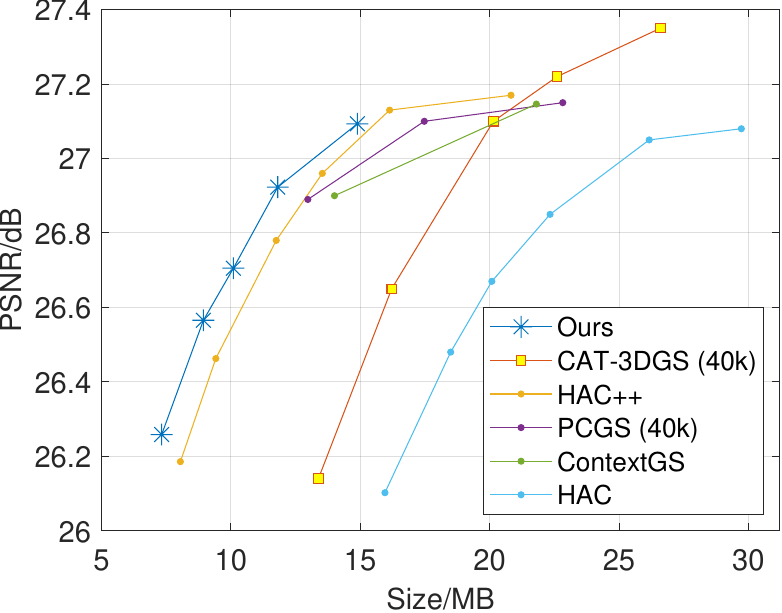}
        \caption{BungeeNeRF}
    \end{subfigure}
    
    \caption{Comparison of our method (HAC+SHTC) with anchor-based methods on two additional datasets, Synthetic-NeRF~\cite{mildenhall2021nerf} and BungeeNeRF~\cite{xiangli2022bungeenerf}.}
    \label{fig:curves_synthetic_bungee_nerf}
\end{figure}

\begin{figure}
  \begin{center}
    \centerline{\includegraphics[width=0.5\columnwidth]{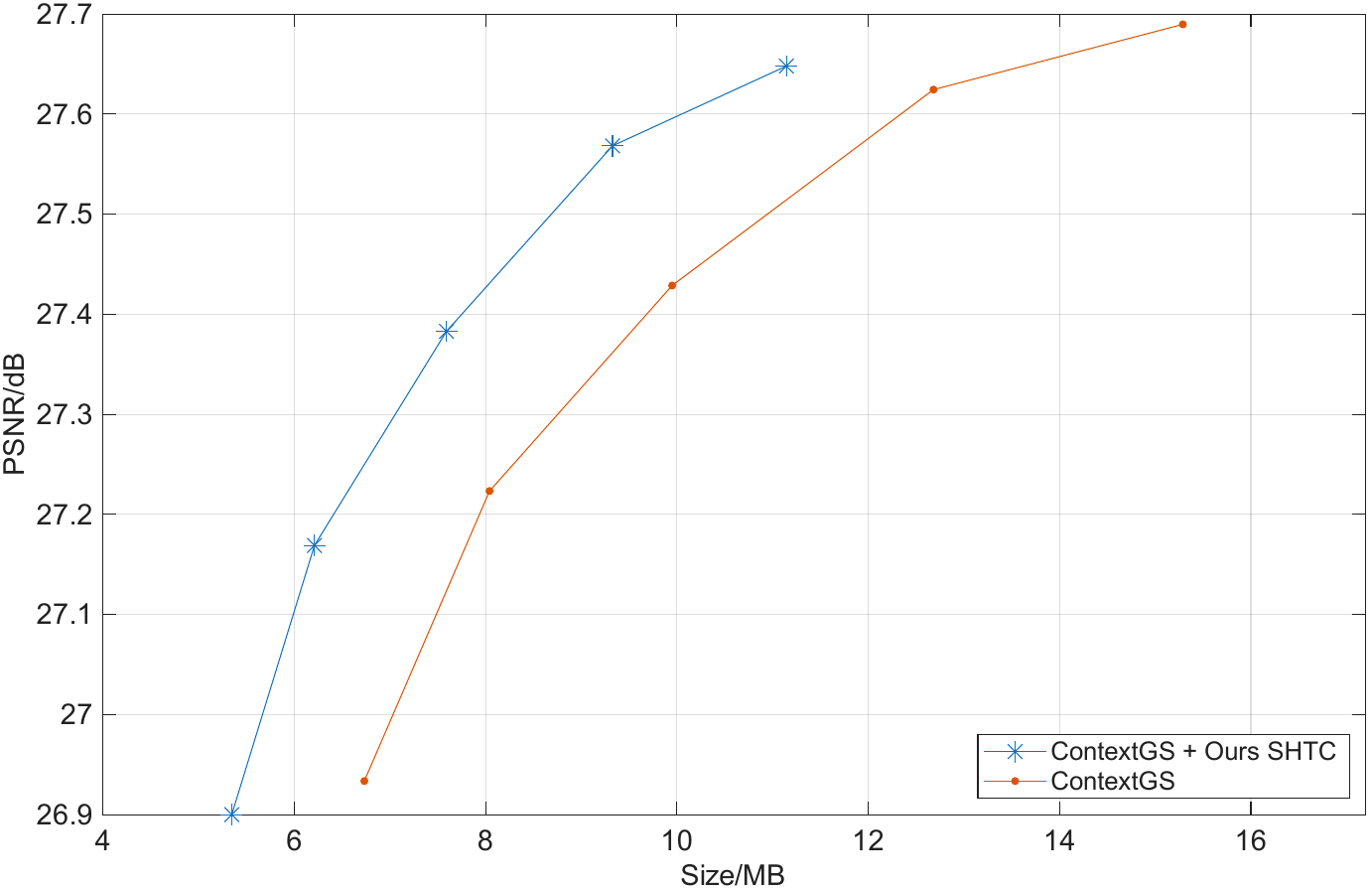}}
    \caption{
    R-D curves of ContextGS with and without SHTC. Integrating SHTC into ContextGS consistently improves compression efficiency, achieving a BD-rate of $-19.45\%$ relative to vanilla ContextGS.
    }
    \label{fig:context_gs_shtc}
  \end{center}
\end{figure}

\begin{figure}
    \centering
    \includegraphics[width=0.9\linewidth]{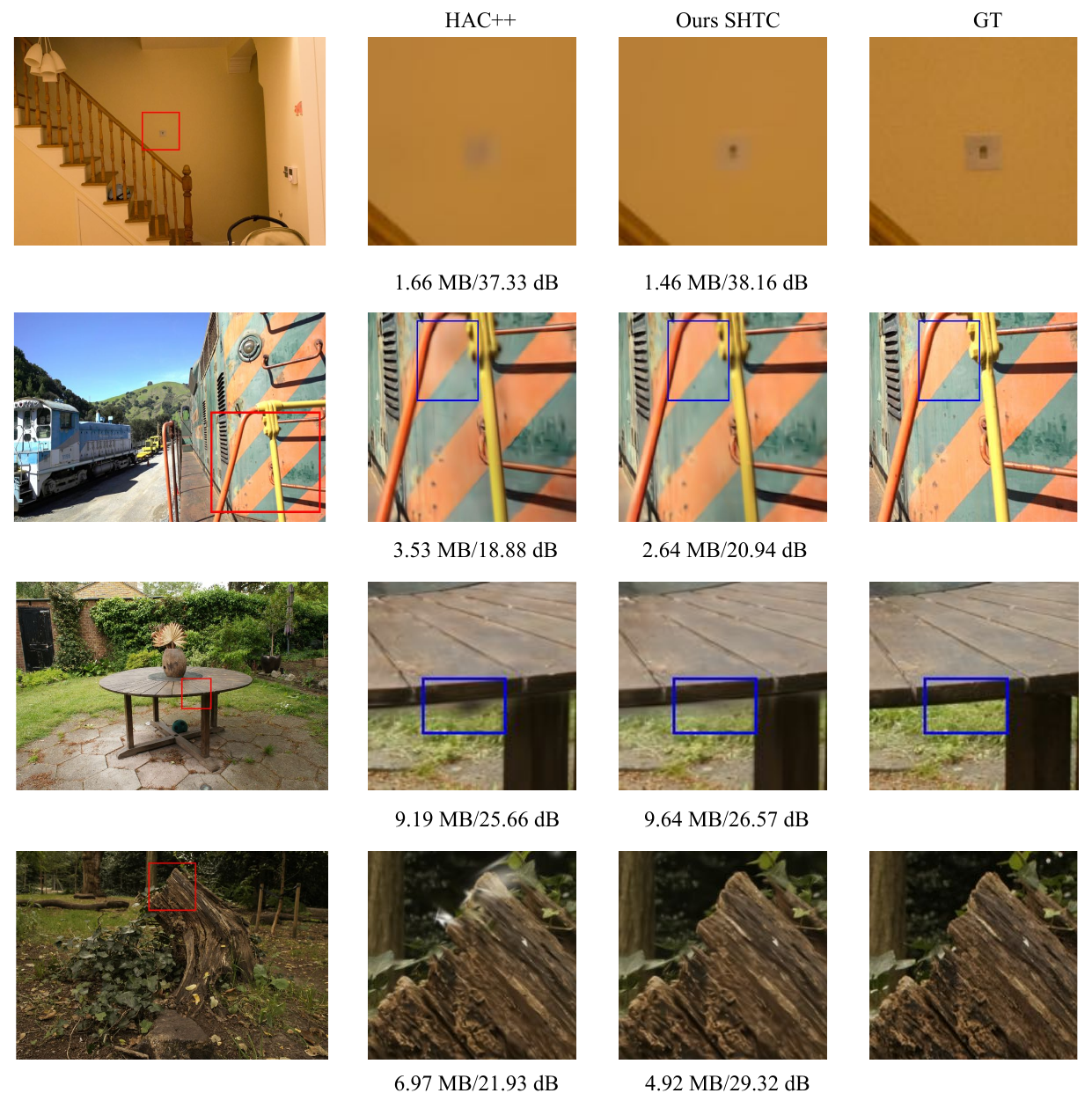}
    \caption{Visual comparison between HAC++ and our SHTC method on four scenes: `playroom' (DeepBlending), `train' (Tanks \& Temples), `garden' and `stump' (Mip-NeRF360). }
    \label{fig:SHTC_visual}
\end{figure}

\section{Visual Comparison}
\label{sec:add_visual}
We provide visual comparisons between our SHTC approach and HAC++~\cite{chen2025hac++} in \cref{fig:SHTC_visual}. For each example, we report the bitstream size (in MB) and the PSNR (in dB) of the shown crop. To illustrate how SHTC reduces rendering distortion under a smaller or comparable memory footprint, we evaluate four representative scenes: `playroom' (DeepBlending), `train' (Tanks \& Temples), and `garden' and `stump' (Mip-NeRF360), which are shown from the first to the fourth row in \cref{fig:SHTC_visual}. In the `playroom' scene, our method better preserves the structure of the wall-mounted switch, which appears as an unrecognizable blur in the HAC++ reconstruction. In the `train' scene, our method suppresses shading and dark banding artifacts along the painted stripes. In the `garden' scene, our method removes the black floater near the table edge. In the `stump' scene, our method effectively reduces floater artifacts around the tree stump.

\begin{figure*}[t]
    \centering    
    \begin{subfigure}[b]{0.45\textwidth}
        \includegraphics[width=0.9\linewidth]{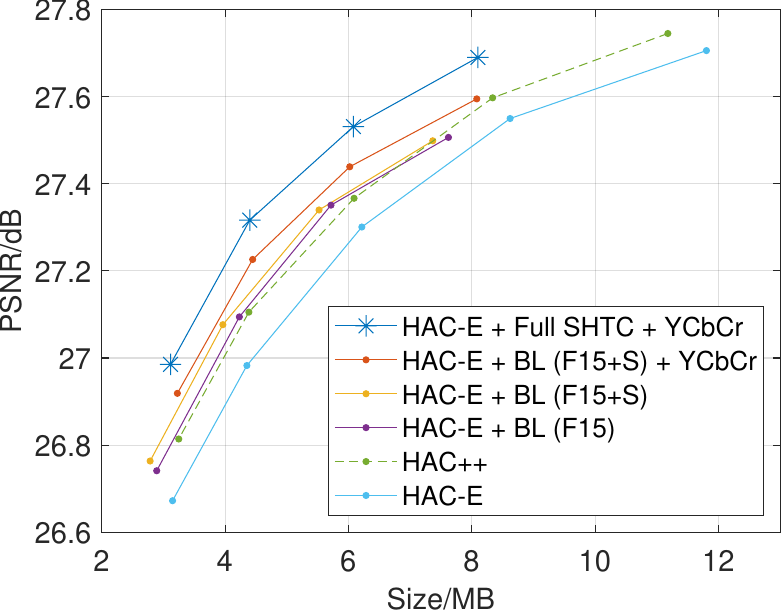}
        \caption{R-D curves for HAC-E, HAC++, and SHTC variants obtained by progressively adding the KLT-based base layer, scaling transform, YCbCr-based distortion, and the full two-layer SHTC.}
        \label{fig:ablation_mip_360_all_variants}
    \end{subfigure}
    \begin{subfigure}[b]{0.45\textwidth}
        \includegraphics[width=0.9\linewidth]{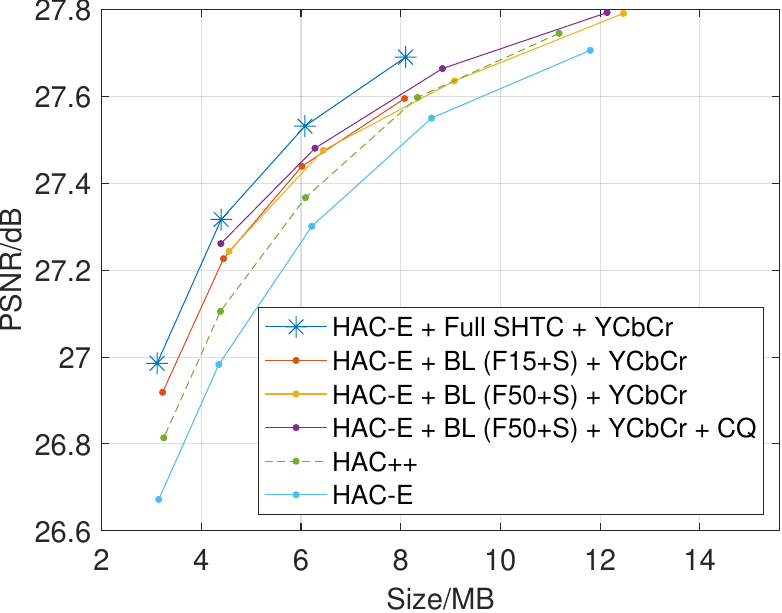}
        \caption{Comparison between our truncated-plus-refinement scheme (HAC-E + Full SHTC + YCbCr) and two variants that transmit all KLT coefficients without truncation.}
        \label{fig:ablation_mip_360_no_truction}
    \end{subfigure}
    \caption{Ablation Study of SHTC Variants on Mip-NeRF360.}
    \label{fig:curves_mip360_ablation_study}
\end{figure*}

\section{Additional Ablation Study}
\label{sec:additional_ablation_study}

\subsection{Additional Quantitative Ablation Study on Component Contributions}
\label{sec:additional_ablation_study_component}
In the main paper, we only report the most basic configuration and the complete version of our method, mainly to let readers grasp the overall benefit of SHTC at a glance under the page limit. In this subsection, we present a more fine-grained quantitative ablation study to isolate and measure the contribution of each component that bridges the basic configuration and the full model.

To more accurately quantify the gains brought by the proposed transform modules, we construct an enhanced version of HAC, denoted by \textbf{HAC-E}. HAC-E is an intermediate variant between the vanilla HAC and HAC++. Following HAC++, HAC-E uses MPEG-GPCC to compress anchor positions and adopts the mask-aware rate estimator, but it still retains the simple entropy model of HAC and does not use complex context modeling. Starting from this common baseline, there are two ways to further improve compression performance:
\begin{enumerate}
    \item replacing the simple entropy model with a highly complex one, which leads to HAC++;
    \item keeping the simple, fast entropy model while introducing learned transforms to enhance R-D performance, which is our transform-coding paradigm.
\end{enumerate}

Along the second route, we build several SHTC variants on top of HAC-E, and their BD-rate values over HAC-E directly quantify the gains of each component. Starting from HAC-E, we progressively add components:
\begin{itemize}
    \item \textbf{HAC-E + BL(F15)}: we activate the most basic version of SHTC and use only its base layer (BL), implemented as a KLT, to compress anchor features. The dimensionality of the principal coefficients is set to 15.
    \item \textbf{HAC-E + BL(F15+S)}: we use the base layer to compress both anchor features and anchor scalings, as described in the main text.
    \item \textbf{HAC-E + BL(F15+S) + YCbCr}: building on the previous variant, we replace the original pixel-wise $\ell_1$ distortion with the proposed YCbCr-space distortion $\mathcal{L}_{\mathrm{YCbCr}}$ in the training objective.
    \item \textbf{HAC-E + Full SHTC + YCbCr}: finally, we use the full two-layer SHTC for anchor feature compression, so that the neural refinement layer compensates for truncation loss at low rate overhead, while the base layer is still used for anchor scaling compression. This corresponds to our full model.
\end{itemize}

We plot the R--D curves of these variants on the Mip-NeRF360 dataset in \cref{fig:ablation_mip_360_all_variants}. Mip-NeRF360 contains diverse and challenging indoor and outdoor scenes, making it a suitable benchmark for ablation with reduced bias. From \cref{fig:ablation_mip_360_all_variants}, we observe that introducing a KLT-based base layer for anchor feature compression (HAC-E + BL(F15)) already yields a clear gain over the HAC-E baseline and brings the performance close to HAC++. Extending the base layer to also compress anchor scalings (HAC-E + BL(F15+S)) and further switching to the YCbCr-based distortion metric (HAC-E + BL(F15+S) + YCbCr) provide additional, consistent improvements. The complete SHTC configuration trained with the YCbCr-based distortion (HAC-E + Full SHTC + YCbCr) achieves the best overall R-D trade-off among all variants.

\cref{tab:ablation_mip360} quantifies these effects by reporting BD-rate values over HAC-E on Mip-NeRF360. HAC++ is also included to facilitate a direct comparison between the two paradigms: using learned transforms with a simple entropy model versus relying solely on a highly complex entropy model. The results show that even the simplest transform-based variant (HAC-E + BL(F15)) already surpasses HAC++, and each additional component yields further BD-rate reductions, leading to a substantial $29.33\%$ reduction for HAC-E + Full SHTC + YCbCr.

\begin{table}[htbp]
    \centering
    \caption{BD-rate over HAC-E on the Mip-NeRF360 dataset for HAC++ and the proposed SHTC variants. Negative values indicate rate savings over HAC-E, and more negative values correspond to better R-D performance.}
    \footnotesize
    \setlength{\tabcolsep}{3pt}
    \begin{tabular}{lc}
        \toprule
         &BD-rate over HAC-E\\
         \midrule
         HAC++&-10.93\%\\
         \hline
         HAC-E+BL(F15)&-13.30\%\\
         HAC-E+BL(F15+S)&-16.59\%\\
         HAC-E+BL(F15+S)+YCbCr&-20.07\%\\
         HAC-E+Full SHTC+YCbCr&-29.33\%\\
         \bottomrule
    \end{tabular}
    
    \label{tab:ablation_mip360}
\end{table}

\subsection{Comparison with No-Truncation Variants}
\label{sec:additional_ablation_study_no_truncation}
Finally, we investigate whether directly transmitting all KLT coefficients can serve as a competitive alternative to our truncated-plus-refinement design. In \cref{fig:ablation_mip_360_no_truction}, we first consider a variant that does \emph{not} truncate the KLT coefficients of anchor features, denoted by HAC-E + BL(F50+S) + YCbCr. This configuration directly quantizes and entropy-codes all 50 KLT coefficients. Under the same rate-control parameter $\lambda$, it attains higher PSNR than the truncated-only baseline, but at a significantly higher bitrate, so it offers no improvement in overall R-D performance compared with simply truncating the coefficients.

Building on HAC-E + BL(F50+S) + YCbCr, we then introduce a monotonically increasing, channel-wise quantization schedule for the 50-dimensional KLT coefficients,
\begin{equation}
    q_s^{(i)} = q_s \exp(\alpha i),\quad i \in \{0,\ldots,49\},
    \label{eq:quantization_step}
\end{equation}
where $\alpha$ is a learnable scalar and $q_s$ is the per-anchor base quantization step predicted by HAC. The idea is to reduce the bitrate cost of transmitting all coefficients by assigning larger quantization steps to less important channels. This variant, denoted by HAC-E + BL(F50+S) + YCbCr + CQ, yields a modest improvement over HAC-E + BL(F50+S) + YCbCr, but still falls short of our truncated-plus-refinement scheme in terms of R-D performance.

These results confirm that directly transmitting all KLT coefficients, even with a more sophisticated quantization schedule, is rate-inefficient and does not bring meaningful gains in R-D performance. In contrast, truncating the transform coefficients and compensating the lost information via a learned refinement layer provides a more effective and bitrate-efficient way to improve compression performance.

\begin{figure*}[htbp]
    \centering
    \begin{subfigure}[b]{0.32\textwidth}
        \includegraphics[width=\linewidth]{Fig/correlation_energy/hac_playroom_correlation.png}
        \caption{HAC}
    \end{subfigure}
    \begin{subfigure}[b]{0.32\textwidth}
        \includegraphics[width=\linewidth]{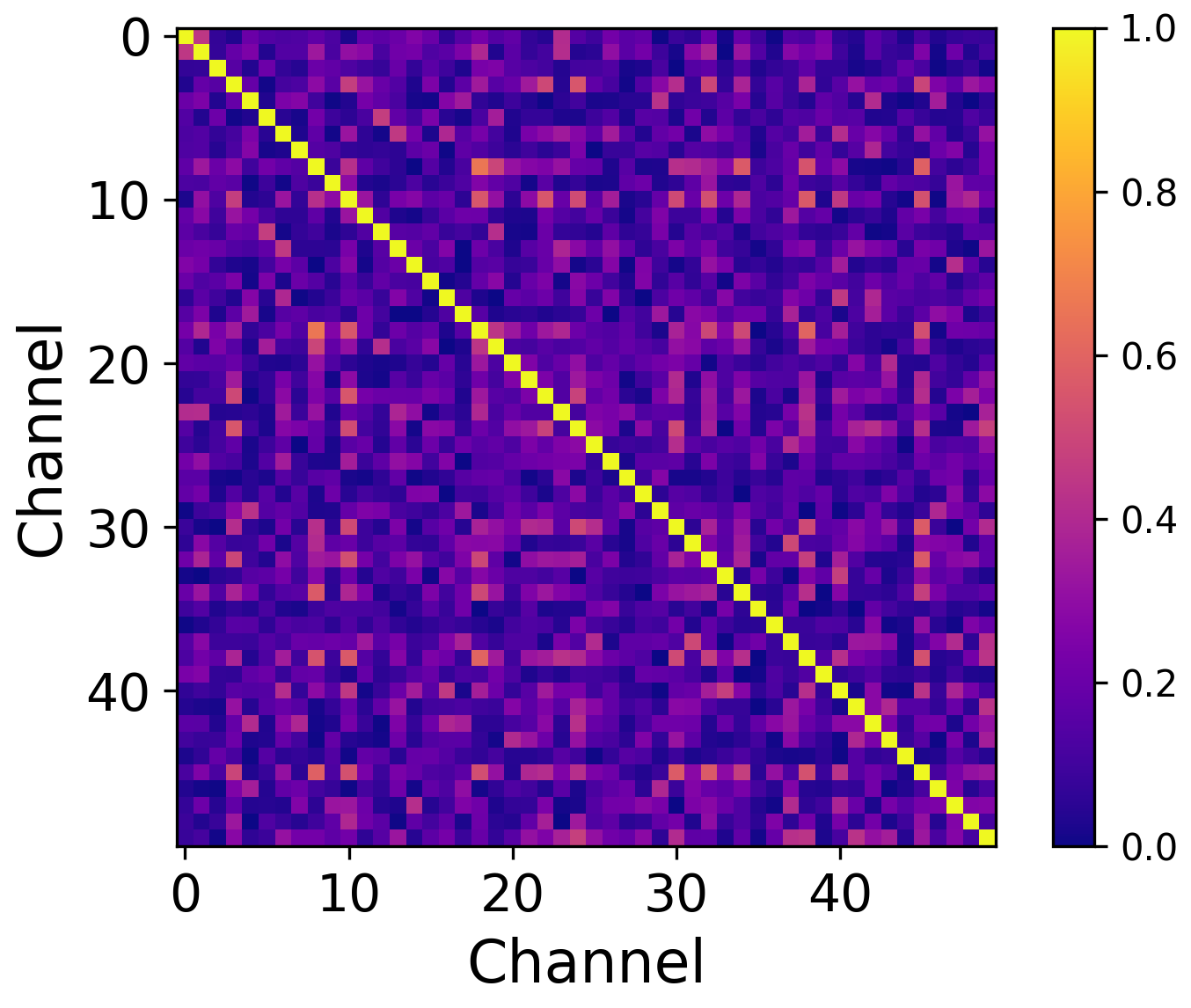}
        \caption{DCT coefficients}
    \end{subfigure}
    \begin{subfigure}[b]{0.32\textwidth}
        \includegraphics[width=\linewidth]{Fig/correlation_energy/playroom_correlation_t.png}
        \caption{KLT coefficients}
    \end{subfigure}
  
    \begin{subfigure}[b]{0.32\textwidth}
        \includegraphics[width=\linewidth]{Fig/correlation_energy/hac_playroom_energy.png}
        \caption{HAC}
    \end{subfigure}
    \begin{subfigure}[b]{0.32\textwidth}
        \includegraphics[width=\linewidth]{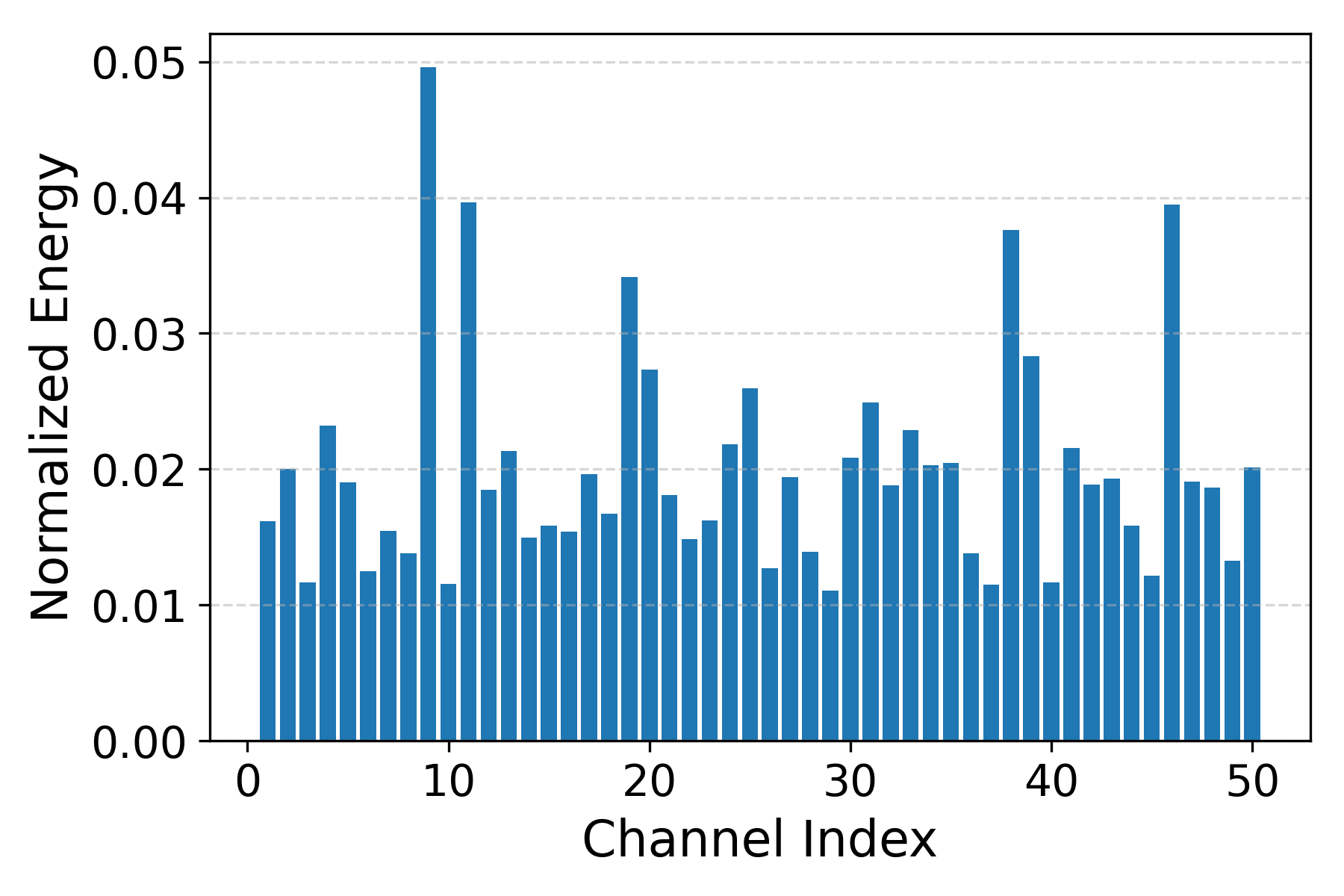}
        \caption{DCT coefficients}
    \end{subfigure}
    \begin{subfigure}[b]{0.32\textwidth}
        \includegraphics[width=\linewidth]{Fig/correlation_energy/playroom_energy_t.png}
        \caption{KLT coefficients}
    \end{subfigure}

    \begin{subfigure}[b]{0.23\textwidth}
        \includegraphics[width=\linewidth]{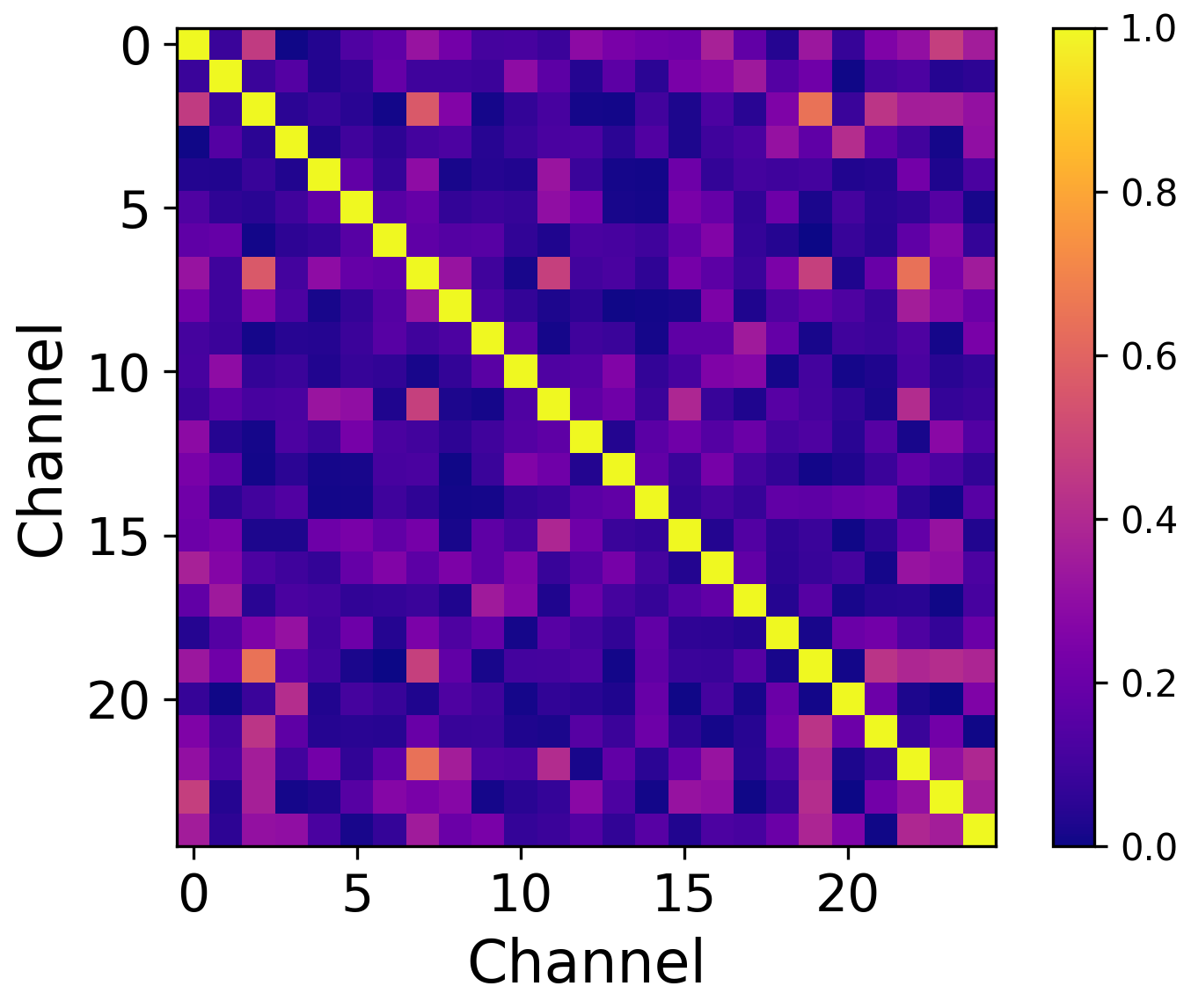}
        \caption{Haar L-components}
    \end{subfigure}
    \begin{subfigure}[b]{0.23\textwidth}
        \includegraphics[width=\linewidth]{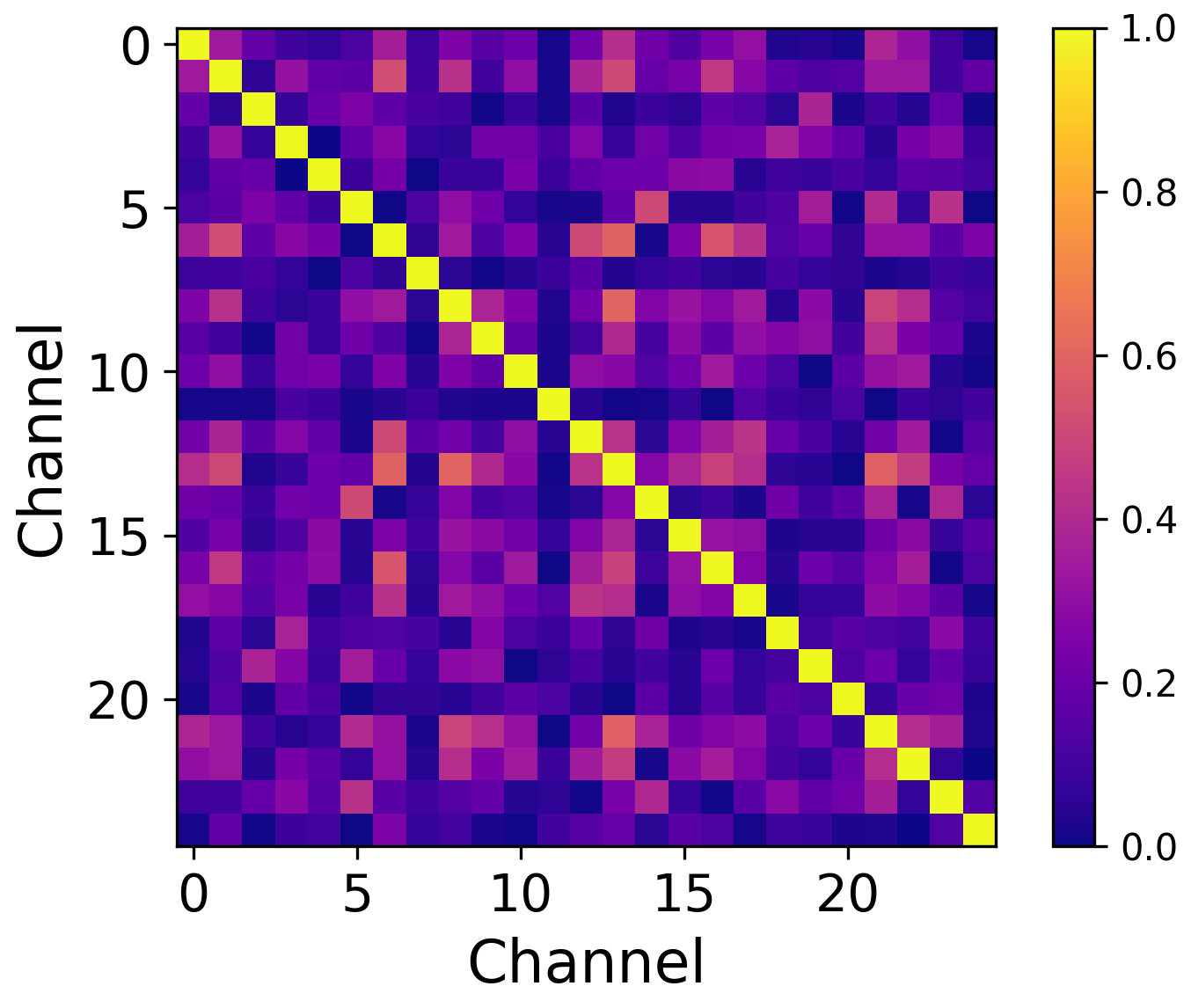}
        \caption{Haar H-components}
    \end{subfigure}
    \begin{subfigure}[b]{0.23\textwidth}
        \includegraphics[width=\linewidth]{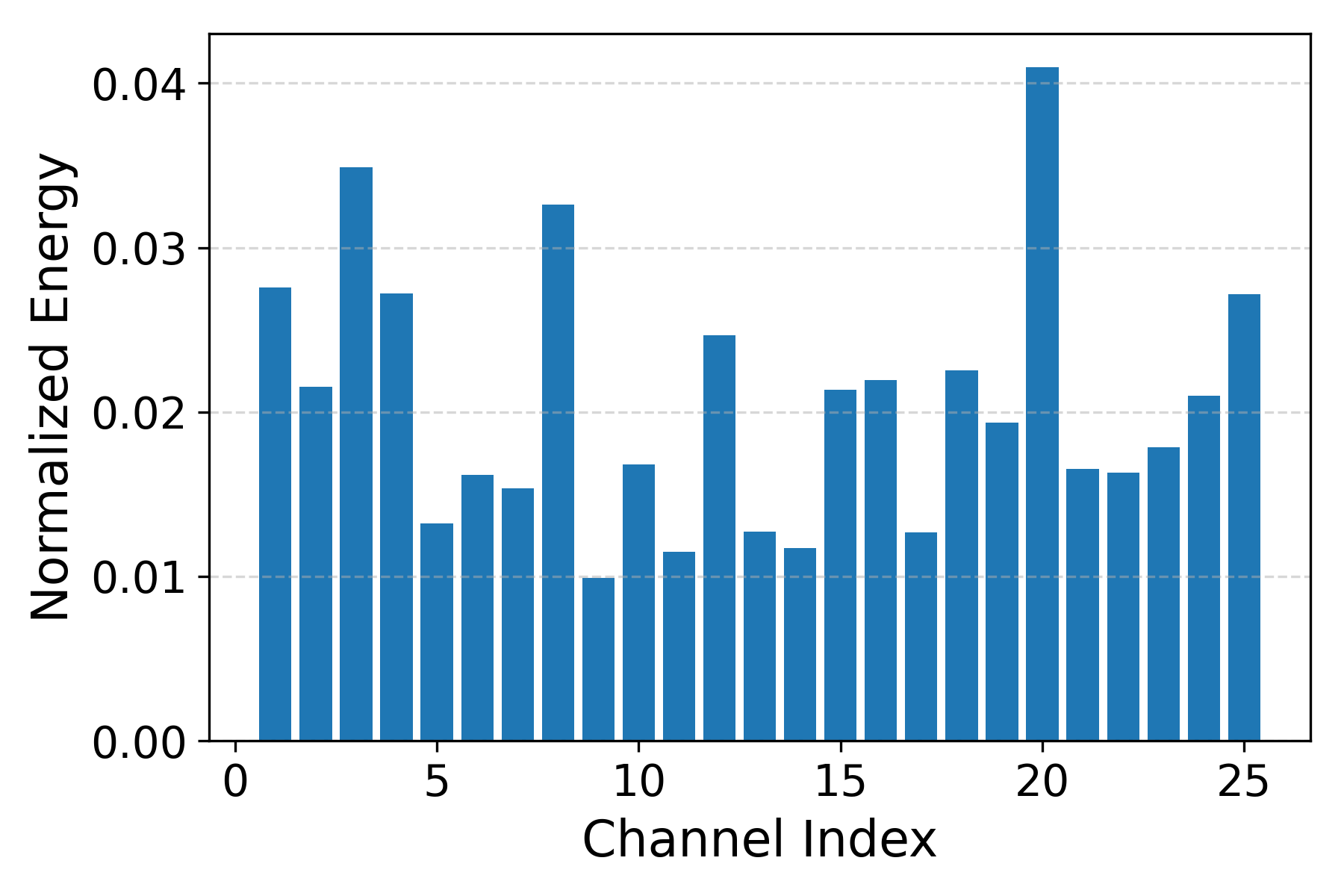}
        \caption{Haar L-components}
    \end{subfigure}
    \begin{subfigure}[b]{0.23\textwidth}
        \includegraphics[width=\linewidth]{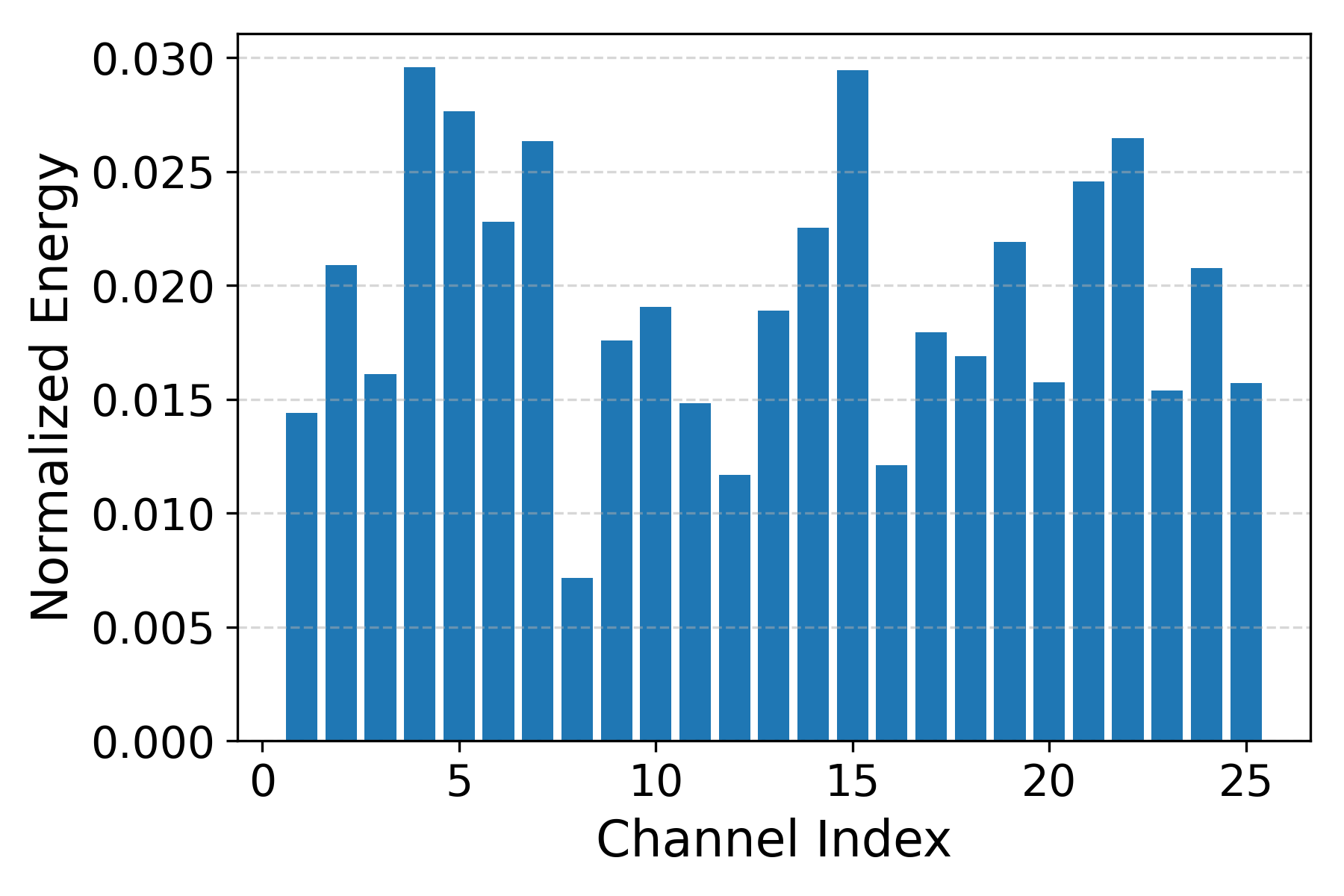}
        \caption{Haar H-components}
    \end{subfigure}
    \caption{Inter-channel decorrelation and energy compaction across transforms. (a-c) Absolute inter-channel correlation matrices for the original data, DCT coefficients, and KLT coefficients. (d–f) Normalized energy per channel. (g-h) Absolute inter-channel correlation for the Haar transform’s two sub-bands. (i–j) Corresponding normalized energy distributions for Haar L and Haar H sub-bands.}
    \label{fig:transform_choices}
\end{figure*}
\section{Discussion of Transform Choices}
\label{sec:transform_choices}
\subsection{Transform Choices for the Base Layer}
\begin{figure}[t]
    \centering    
    \begin{subfigure}[b]{0.45\textwidth}
        \includegraphics[width=0.9\linewidth]{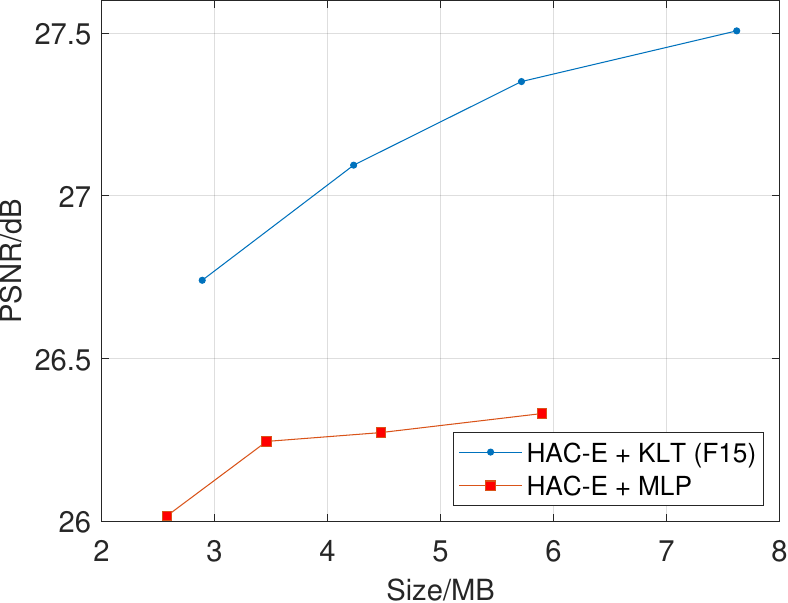}
        \caption{Comparison between `HAC-E + KLT (F15)' and `HAC-E + MLP' on the Mip-NeRF360 dataset.}
    \label{fig:abla_wiht_mlp}
    \end{subfigure}
    \begin{subfigure}[b]{0.45\textwidth}
        \includegraphics[width=0.9\columnwidth]{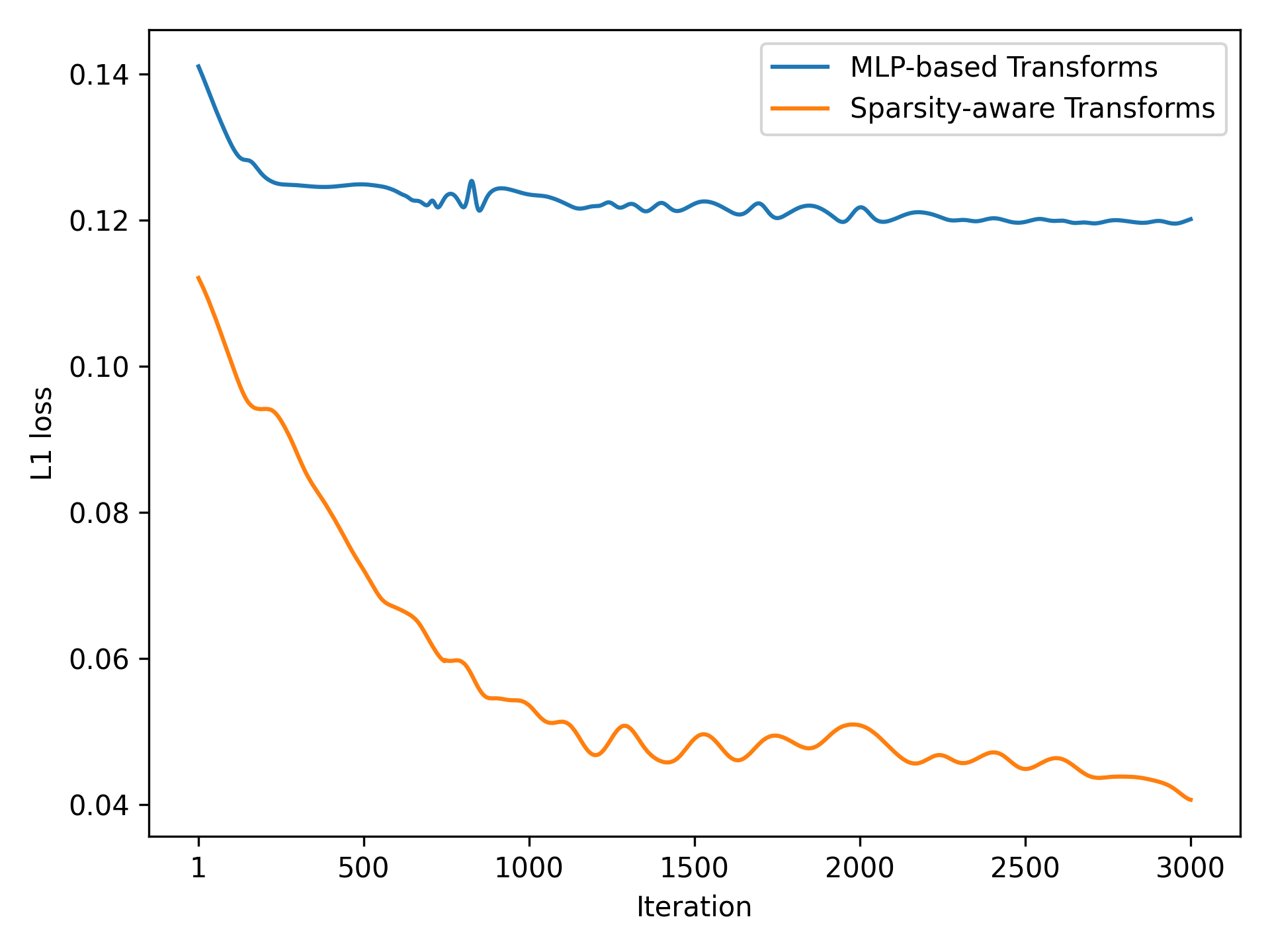}
        \caption{Offline comparison of our sparsity-aware refinement layer and a black-box MLP-based transform on the \textit{bicycle} scene of the Mip-NeRF360 dataset. The plot shows the $\ell_1$ reconstruction loss of the KLT residual versus training iteration.}
    \label{fig:abla_sparsity_aware}
    \end{subfigure}
    \caption{Comparison of different design choices.}
\end{figure}

In this section, we present a high-level comparison of transform design choices, supported by several analysis experiments.
\paragraph{Why Not DCT or DWT?}
We considered several candidates for the base-layer transform, including the classical DCT and the Haar wavelet. To guide this choice, we compared their ability to decorrelate anchor channels and compact energy. \cref{fig:transform_choices} shows the absolute inter-channel correlation matrices and the normalized energy distributions of the original HAC features, the DCT coefficients, the Haar coefficients, and the KLT coefficients. DCT and Haar can reduce inter-channel correlation and slightly improve energy compactness, but their gains are limited: the correlation matrices still exhibit noticeable off-diagonal structure and the energy remains relatively spread out across channels. This behavior is expected because their bases are fixed and data-agnostic. In contrast, KLT yields an almost perfectly diagonal correlation matrix and concentrates most of the energy in the first few coefficients. This data-dependent transform is obtained as the eigenvectors of the sample covariance and is theoretically optimal for decorrelation and energy compaction among linear transforms. The main additional cost, relative to DCT and Haar, is storing the KLT basis. Even if the full basis must be transmitted, a 50-dimensional feature vector only requires a $50 \times 50$ matrix (2,500 coefficients), which is negligible compared with the overall bitstream size.
\paragraph{Why Not an MLP-Based Transform?}
Beyond classical transforms, we also compare our `HAC-E + KLT (F15)' variant (the simplest version of SHTC) with a variant equipped with an MLP-based analysis and synthesis transform pair. Both networks are 5-layer MLPs, with a total of 21,930 parameters that must be transmitted as part of the bitstream. This comparison, evaluated on the Mip-NeRF360 dataset and shown in \cref{fig:abla_wiht_mlp}, reveals that the black-box MLP transform performs significantly worse in terms of the R-D trade-off than the simple KLT-based design. A likely reason is that, under a strict parameter budget, the MLP-based transform lacks sufficient expressive power to accurately reconstruct the input, leading to larger reconstruction errors and thus worse R-D performance. 

For these reasons, we adopt KLT as a cost-effective choice for the base-layer transform. KLT provides strong decorrelation and energy compaction at negligible storage cost, while remaining simple and invertible. On top of this, we use a neural residual coder to compensate for the truncation loss of the KLT coefficients, which further improves the R-D performance over using KLT alone.

\subsection{Transform Choices for the Refinement Layer}
When designing the neural refinement layer, a natural question arises: why not simply use a generic MLP? Our key observation is that the residual to be coded by the refinement layer (i.e., the KLT residual) can be reasonably modeled as a sparse signal. Classical compressed sensing theory shows that sparse signals can be recovered from a small number of linear measurements by solving a sparsity-regularized inverse problem. This perspective suggests a principled way to construct the refinement layer: instead of relying on an unconstrained black-box network, we explicitly exploit sparsity and build the module on top of a well-understood sparse reconstruction framework, so that each component has a clear interpretation.

In 3DGS compression, the parameters of transforms must be transmitted as part of the bitstream, so the parameter budget is limited. Under such constraints, a purely neural, black-box MLP may not have sufficient effective capacity to learn a good transform and its approximate inverse from data. In contrast, a sparsity-aware design embeds strong structure and prior knowledge directly into the architecture, so that even with relatively few parameters it can still achieve accurate reconstruction. This leads to much higher parameter efficiency than a generic MLP-based transform.

To verify this intuition, we conduct several offline experiments; here we present results on the \textit{bicycle} scene of the Mip-NeRF360 dataset, where we compress the KLT residual in two different ways. The first variant uses a pure MLP-based transform with an MLP analysis and synthesis transform pair; both networks are 5-layer MLPs, resulting in a total of 21,930 parameters that must be transmitted in the bitstream. The second variant uses our sparsity-aware refinement layer, which contains only 5,093 parameters. We train both models for 3{,}000 iterations and measure the $\ell_1$ reconstruction loss of the KLT residual. As shown in \cref{fig:abla_sparsity_aware}, the sparsity-aware transform consistently drives the loss down to a much lower level than the MLP-based transform, despite using roughly four times fewer parameters. This early-stage verification supports our motivation: leveraging the sparsity prior and using a compressed-sensing-inspired architecture for residual coding yields a highly parameter-efficient refinement module. For this reason, we adopt the sparsity-aware design as our default choice for the neural refinement layer in SHTC.

\begin{table*}[h]
\centering
\caption{Comparison of our SHTC framework with other 3DGS data compression methods, including 3DGS and Scaffold-GS, for reference. The best and second-best results are highlighted in \colorbox{pink}{\textcolor{black}{red}} and \colorbox{yellow!50}{\textcolor{black}{yellow}} cells, respectively. The size values are measured in megabytes (MB).}
\label{tab:main_results}
\setlength{\tabcolsep}{3pt}
\renewcommand{\arraystretch}{1.3}

\begin{adjustbox}{width=\textwidth}
\begin{tabular}{l|cccc|cccc|cccc}
\hline
\rowcolor[HTML]{FFFFFF} 
\textbf{Datasets}  & \multicolumn{4}{c|}{\textbf{Mip-NeRF360}} & \multicolumn{4}{c|}{\textbf{Tank\&Temples}} & \multicolumn{4}{c}{\textbf{DeepBlending}} \\ \hline
\rowcolor[HTML]{FFFFFF} 
\textbf{Methods} & \textbf{psnr$\uparrow$} & \textbf{ssim$\uparrow$} & \textbf{lpips$\downarrow$} & \textbf{size$\downarrow$}  & \textbf{psnr$\uparrow$} & \textbf{ssim$\uparrow$} & \textbf{lpips$\downarrow$} & \textbf{size$\downarrow$}  & \textbf{psnr$\uparrow$} & \textbf{ssim$\uparrow$} & \textbf{lpips$\downarrow$} & \textbf{size$\downarrow$} \\ \hline
3DGS~\cite{kerbl20233d} & 27.46 & \cellcolor{yellow!50}{0.812} & \cellcolor{yellow!50}{0.222} & 750.9 & 23.69 & 0.844 & {0.178} & 431.0 & 29.42 & 0.899 & \cellcolor{yellow!50}{0.247} & 663.9 \\ 
Scaffold-GS~\cite{lu2024scaffold} & 27.50 & 0.806 & 0.252 & 253.9 & 23.96 & \cellcolor{yellow!50}{0.853} & \cellcolor{yellow!50}{0.177} & 86.50 & \cellcolor{pink}{30.21} & \cellcolor{yellow!50}{0.906} & 0.254 & 66.00  \\ 
\hline

Compact3DGS~\cite{lee2024compact} & 27.08 & 0.798 & 0.247 & 48.80 & 23.32 & 0.831 & 0.201 & 39.43 & 29.79 & 0.901 & 0.258 & 43.21 \\ 
Compressed3D~\cite{niedermayr2024compressed} & 26.98 & 0.801 & 0.238 & 28.80 & 23.32 & 0.832 & 0.194 & 17.28 & 29.38 & 0.898 & 0.253 & 25.30  \\ 
EAGLES~\cite{girish2024eagles} & 27.14 & {0.809} & 0.231 & 58.91 & 23.28 & 0.835 & 0.203 & 28.99 & 29.72 & \cellcolor{yellow!50}{0.906} & {0.249} & 52.34  \\ 
LightGaussian~\cite{fan2024lightgaussian} & 27.00 & 0.799 & 0.249 & 44.54 & 22.83 & 0.822 & 0.242 & 22.43 & 27.01 & 0.872 & 0.308 & 33.94  \\ 

Navaneet \textit{et al.}~\cite{navaneet2024compgs} & 27.12 & 0.806 & {0.240} & 19.33 & 23.44 & 0.838 & 0.198 & 12.50 & 29.90 & \cellcolor{pink}{0.907} & {0.251} & 13.50  \\ 
Reduced3DGS~\cite{papantonakis2024reducing} & 27.19 & 0.807 & {0.230} & 29.54 & 23.57 & 0.840 & 0.188 & 14.00 & 29.63 & 0.902 & {0.249} & 18.00  \\ 
RDOGaussian~\cite{wang2024end} & 27.05 & 0.802 & {0.239} & 23.46 & 23.34 & 0.835 & 0.195 & 12.03 & 29.63 & 0.902 & {0.252} & 18.00  \\ 
PUP 3D-GS~\cite{hanson2025pup}& 26.67 & 0.786 & 0.272 & 74.65 & 22.72 & 0.801 & 0.244 & 43.33 & 28.85 & 0.881 & 0.302 & 69.92  \\ 
SizeGS~\cite{xie2025sizegs}& 27.48 & 0.806 & 0.240 & 18.17 & 24.04 & 0.840 & 0.200 & 10.93 & 30.24 & 0.903 & 0.271 & 7.92  \\
FlexGaussian~\cite{tian2025flexgaussian}& 26.38 & 0.780 & 0.251 & 40.80 & 22.44 & 0.804 & 0.219 & 16.30 & 28.61 & 0.884 & 0.269 & 25.48  \\
\hline
NeuralGS~\cite{tang2025neuralgs}& 27.35 & 0.806 & 0.240 & 16.90 & 23.63 & 0.841 & 0.192 & 12.06 & 29.91 & \cellcolor{yellow!50}{0.906} & 0.254 & 12.98  \\
LocoGS~\cite{shin2025localityaware} & 27.40 & \cellcolor{pink}{0.815} & \cellcolor{pink}{0.219}& 13.89 & 23.89 & \cellcolor{pink}{0.854} & \cellcolor{pink}{0.160} & 12.34 & 30.17 & \cellcolor{yellow!50}{0.906} & \cellcolor{pink}{0.244} & 13.39  
\\
\hline
FCGS (low rate)~\cite{chen2025fast} & 27.05 & 0.798 & 0.237 & 36.30 & 23.48 & 0.833 & 0.193 & 18.80 & 29.27 & 0.893 & 0.257 & 30.10 
\\
FCGS (high rate)~\cite{chen2025fast} & 27.39 & 0.806 & {0.226}& 67.20 & 23.62 & 0.839 & 0.184 & 33.60 & 29.58 & 0.899 & {0.248} & 54.50  
\\
\hline
SOG~\cite{morgenstern2024compact} & 26.56 & 0.791 & 0.241 & 16.70 & 23.15 & 0.828 & 0.198 & 9.30 & 29.12 & 0.892 & 0.270 & 5.70  \\ 
CodecGS~\cite{lee2025compression} & 27.30 & {0.810} & 0.236 & 9.78 & 23.63 & 0.842 & 0.192 & 7.46 & 29.82 & \cellcolor{pink}{0.907} & 0.251 & 8.62  \\ 
MesonGS~\cite{xie2024mesongs} & 26.99 & 0.796 & 0.247 & 27.16 & 23.32 & 0.837 & 0.193 & 16.99 & 29.51 & 0.901 & 0.251 & 24.76  \\ 
\hline
CompGS~\cite{liu2024compgs}&27.26&0.803&0.239&16.50&23.70&0.837&0.208&9.60&29.69&0.901&0.279&8.77\\
FP-Net~\cite{ma2025enhancing}& 27.46 & 0.801 & 0.249 & 11.08 & 24.13 & 0.847 & 0.192 & 6.34 & 30.14 & \cellcolor{yellow!50}{0.906} & 0.274 & 3.46  \\
FP-GS~\cite{tang2025feature}& 27.53 & 0.806 & 0.240 & 12.82 & \cellcolor{yellow!50}{24.32} & 0.811 & 0.185 & 7.32 & 30.05 & \cellcolor{yellow!50}{0.906} & 0.266 & 3.91  \\
HAC~\cite{chen2024hac}  & 27.53 &0.807 & 0.238 & 15.26 & {24.04} & 0.846 & {0.187} & 8.10 & 29.98 & 0.902 & 0.269 &  {4.35} 
\\ 
Context-GS~\cite{wang2024contextgs} & \cellcolor{yellow!50}{27.62} & {0.808} & 0.237 & 12.68 & {24.20} & {0.852} & {0.184} & 7.05 & {30.11} & \cellcolor{pink}{0.907} & 0.258 & 3.45
\\
HAC++~\cite{chen2025hac++}  &  {27.60} & {0.803} & 0.253 & \cellcolor{yellow!50}{8.34} & {24.22} & 0.849 & {0.190} & \cellcolor{pink}{5.18} & {30.16} & \cellcolor{pink}{0.907} & 0.266 &  \cellcolor{yellow!50}{2.91} 
\\
\hline
Ours& \cellcolor{pink}{27.69} & 0.804 & 0.248 & \cellcolor{pink}{8.10} & \cellcolor{pink}{24.34} & 0.845 & 0.191 & \cellcolor{yellow!50}{5.19} & \cellcolor{yellow!50}{30.19} & {0.903} & 0.272 & \cellcolor{pink}{2.49} 
\\ 
\hline
\end{tabular}
\end{adjustbox}

\end{table*}

\begin{figure*}[t]
    \centering    
    \begin{subfigure}[b]{0.32\textwidth}
        \includegraphics[width=\linewidth]{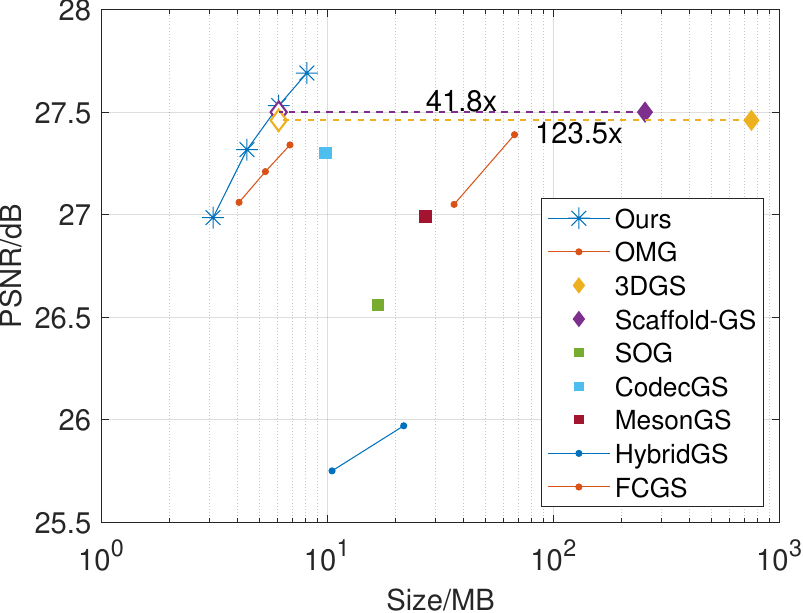}
        \caption{Mip-NeRF360}
    \end{subfigure}
    \begin{subfigure}[b]{0.32\textwidth}
        \includegraphics[width=\linewidth]{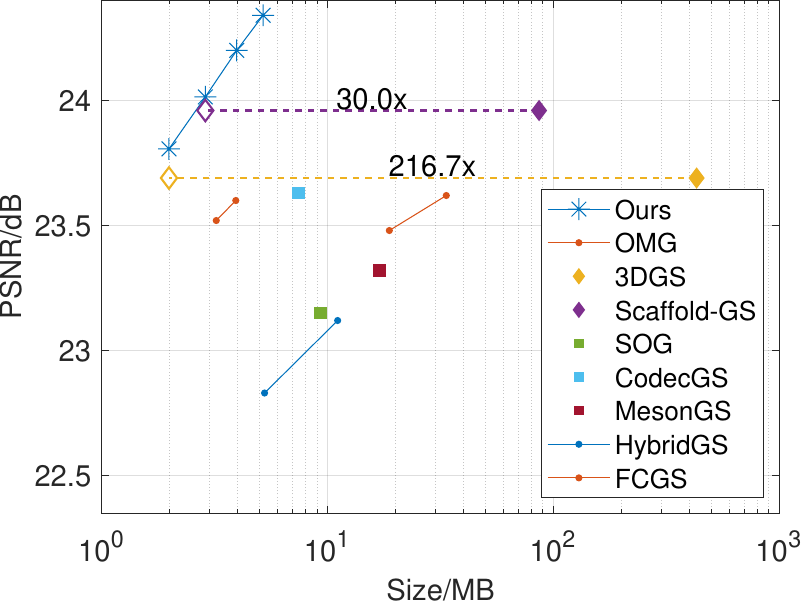}
        \caption{Tank\&Temples}
    \end{subfigure}
    \begin{subfigure}[b]{0.32\textwidth}
        \includegraphics[width=\linewidth]{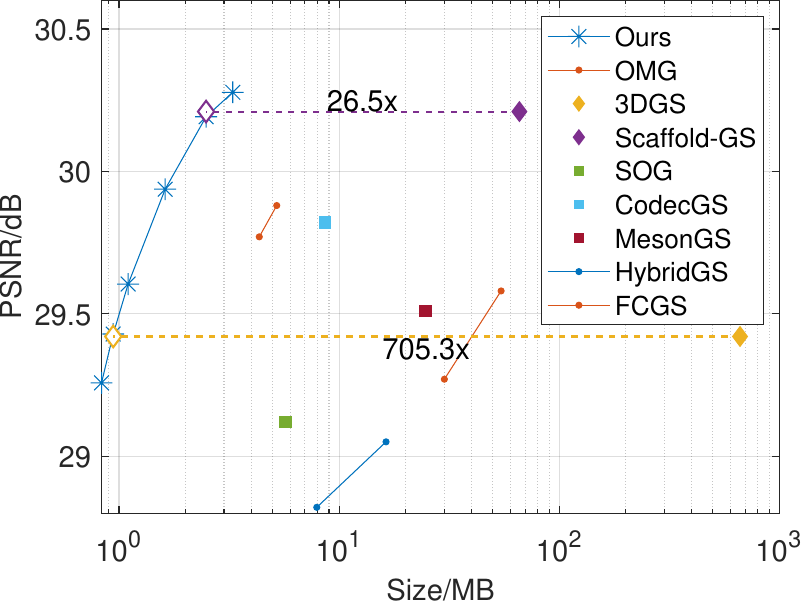}
        \caption{DeepBlending}
    \end{subfigure}
    \caption{Comparison of our method with several conceptually representative baslines.}
    \label{fig:curves_other_methods}
\end{figure*}
\section{Additional Comparisons with Other Methods}
\label{sec:other_method}
In addition to the comparisons with anchor-based compression methods using R–D curves and BD-rate, we also evaluate our method against a broad set of anchor-free and hybrid baselines. These include several commonly used 3DGS and compression baselines~\cite{kerbl20233d,lu2024scaffold,lee2024compact,niedermayr2024compressed,girish2024eagles,fan2024lightgaussian,morgenstern2024compact,navaneet2024compgs,papantonakis2024reducing,wang2024end,liu2024compgs}, the feed-forward compression method FCGS~\cite{chen2025fast}, and two neural-field-based 3DGS compression methods, LocoGS~\cite{shin2025localityaware} and NeuralGS~\cite{tang2025neuralgs}. We further consider several recent 3DGS compression methods~\cite{hanson2025pup,xie2025sizegs,tian2025flexgaussian,lee2025optimized}, two HAC-based approaches that introduce prediction modules to improve HAC~\cite{tang2025feature,ma2025enhancing}, and a group of methods that apply transform coding in a post-training stage for 3DGS compression~\cite{morgenstern2024compact,lee2025compression,xie2024mesongs,yang2025hybridgs}. 

Regarding SOGS~\cite{zhang2025sogs}, we do not include it in our quantitative comparison for two reasons. First, to the best of our knowledge, there is no official implementation available, and the original paper does not report file sizes or distortion metrics on the datasets used in our experiments, which makes a fair and reproducible comparison difficult. Second, SOGS is essentially a 3DGS compaction method rather than a full compression scheme with an explicit bitstream and entropy coding. Based on the model statistics reported in~\cite{zhang2025sogs}, we can only roughly infer an effective compression ratio of about $2\times$ over Scaffold-GS; this is merely a rough estimate, since the authors do not provide explicit compression ratios. In contrast, by employing the proposed SHTC framework, our method achieves substantially higher compression ratios over Scaffold-GS. For these reasons, we omit SOGS from the quantitative comparison.

Because many of the above methods report only one or two operating points on the R–D curve, a meaningful BD-rate computation is not feasible. For these methods, we therefore provide a numerical comparison in tabular form in \cref{tab:main_results}. As shown in \cref{tab:main_results}, our method achieves a better memory-quality trade-off than above approaches.

Averaged over the three standard datasets in Fig.~\ref{fig:curves_other_methods}, our method reduces the memory footprint by about 349$\times$ relative to vanilla 3DGS and by about 33$\times$ relative to Scaffold-GS, while maintaining comparable rendering quality.

\begin{table*}
    \centering
    \caption{Summary of $\lambda$ values used to generate R-D curves on different datasets.}
    \label{tab:hyperparamter_lmbda_summary}
    \begin{adjustbox}{width=\linewidth}
    \begin{tabular}{lccccc}
    \toprule
    &HAC&ContextGS&HAC++&Ours\\
    \midrule
    Mip-NeRF360& $\{0.002,0.004,0.008,0.015\}$& $\{0.002,0.004,0.008,0.015,0.025\}$& $\{0.002,0.004,0.008,0.015,0.025\}$&$\{0.002,0.004,0.008,0.015\}$\\
    Tank\&Temples&$\{0.002,0.004,0.008,0.015,0.025\}$&$\{0.002,0.004,0.008,0.015,0.025,0.04\}$&$\{0.002,0.004,0.008,0.015,0.025\}$&$\{0.002,0.004,0.008,0.015\}$\\
    DeepBlending&$\{0.002,0.004,0.008,0.015,0.025\}$&$\{0.002,0.004,0.008,0.015,0.025,0.04\}$&$\{0.002,0.004,0.008,0.015,0.025,0.04\}$&$\{0.002,0.004,0.008,0.015, 0.02, 0.025\}$\\
    Synthetic-NeRF&$\{0.0005,0.001,0.002,0.003,0.004\}$&-&$\{0.0005,0.001,0.002,0.003,0.004\}$&$\{0.0005,0.001,0.002,0.003,0.004\}$\\
    BungeeNeRF&$\{0.001,0.002,0.003,0.004,0.006\}$&$\{0.001,0.004\}$&$\{0.0005,0.001,0.002,0.003,0.004,0.006,0.008\}$&$\{0.001,0.002,0.003,0.004,0.006\}$\\
    \bottomrule
    \end{tabular}
    \end{adjustbox}
    
\end{table*}

\section{Implementation Details}
\label{sec:implementation_details}
The overall training loss is defined as
\begin{equation}
    \mathcal{L}=\mathcal{L}_{d}+\lambda\mathcal{L}_{\mathrm{b-rate}}+\lambda_o\mathcal{L}_{\mathrm{reg}}+\lambda_e\ell_1(\mathbf{r},\hat{\mathbf{r}})+\lambda_r\mathcal{L}_{\mathrm{r-rate}}
    \label{eq:loss_function}
\end{equation}
where $\mathcal{L}_{d}$ denotes the rendering distortion. 
In practice, $\mathcal{L}_{d}$ is implemented as a weighted sum of  $\mathcal{L}_{\mathrm{YCbCr}}$ and $1-\mathrm{SSIM}(I,\hat{I})$. Details on $\mathcal{L}_{\mathrm{YCbCr}}$ are provided in Appendix~\ref{sec:distortion_term}.
The term $\mathcal{L}_{\mathrm{b-rate}}$ represents the estimated bit rate of $\boldsymbol{\theta}_p$, $\mathbf{l}_t$, and $\{\mathbf{O}_i\}_{i=1}^{K}$, while $\mathcal{L}_{\mathrm{r-rate}}$ represents the rate of $\mathbf{y}$. The regularization term $\mathcal{L}_{\mathrm{reg}}$ includes the additional loss terms inherited from the HAC framework, and $\ell_1(\mathbf{r},\hat{\mathbf{r}})$ encourages accurate reconstruction of the residual signal. When estimating the rate during training, we follow HAC++~\cite{chen2025hac++} and take the anchor mask into account. We set $\lambda_e = 0.03$ and $\lambda_r = \max(\lambda/4, 0.001)$. The setup of $\lambda$ values is provided in Appendix~\ref{sec:lambda_of_curves}. For other hyperparameters not explicitly mentioned, we adopt the default settings consistent with HAC and HAC++.

\subsection{Pixel-wise Error in YCbCr Color Space}
\label{sec:distortion_term}
We propose a new distortion term for supervising the rendered images, inspired by the YCbCr color transform used in traditional image and video codecs. Because humans are more sensitive to luminance variations than to chrominance, we compute pixel-wise $\ell_1$ loss in the YCbCr space and place a higher weight on the Y component. To preserve fine detail, we add a high-frequency fidelity term on the Y channel by encouraging small differences between the Laplacian-filter response maps of $I_Y$ and $\hat{I}_Y$. To suppress color noise and artifacts, we further impose a total variation (TV) regularizer on the Cb and Cr channels. Together, the distortion term is defined as

\begin{align}
    \mathcal{L}_{\mathrm{YCbCr}}
    = \ell_1(I_Y,\hat{I}_Y)
    + \lambda_c \ell_1(I_{Cb},\hat{I}_{Cb})
    + \lambda_c \ell_1(I_{Cr},\hat{I}_{Cr}) 
    +\lambda_H \ell_1(\nabla^2 I_Y,\nabla^2 \hat{I}_Y)
    + \lambda_{TV} \mathrm{TV}(\hat{I}_{Cb})
    + \lambda_{TV} \mathrm{TV}(\hat{I}_{Cr}) 
\end{align}
This distortion term acts as a regularizer on the compressed 3DGS representation, encouraging it to render images with reduced luminance errors at a given bitrate, since luminance errors are perceptually more important than chrominance errors. This, in turn, reduces the overall distortion and empirically improves the R-D performance. In our implementation, we set $\lambda_c$, $\lambda_H$, and $\lambda_{TV}$ to 0.6, 0.15, and 0.1, respectively.

\subsection{Details of $\lambda$ Selection for R-D Curves}
\label{sec:lambda_of_curves}
In this subsection, we summarize the $\lambda$ values used to obtain the R-D curves in our experiments in \cref{tab:hyperparamter_lmbda_summary}. 
For our primary experiments, we set the Lagrange multiplier $\lambda$ to $\{0.002, 0.004, 0.008, 0.015\}$, which yields operating points spanning a wide rate-distortion range and enables a comprehensive comparison. 
To obtain more uniformly spaced operating points along the R--D curve, we use progressively larger gaps between successive $\lambda$ values. This is motivated by the highly nonlinear relationship between $\lambda$ and the achieved rate. In particular, in the low-rate regime, larger changes in $\lambda$ are often necessary to produce clearly separated rates. 
In addition, we discuss several special cases for baselines, detailing how $\lambda$ (or its equivalent control parameter) and the corresponding R--D points are obtained.
\begin{itemize}
    \item \textbf{CAT-3DGS.} For CAT-3DGS~\cite{zhan2025catdgs}, we encountered out-of-memory (OOM) errors during rendering when using the official implementation on our hardware. Consequently, we do not re-train or re-evaluate this method; instead, we directly use the R-D points reported in the original paper as its results in our plots.
    
    \item \textbf{Synthetic-NeRF.} For experiments on the Synthetic-NeRF dataset, in order to obtain a quick verification of our method under comparable settings, we follow the $\lambda$ configurations recommended in the original HAC and HAC++. For the HAC, HAC++, and ContextGS baselines, we directly adopt the R-D numbers reported by their authors. Our own method is evaluated by running our implementation with the same $\lambda$ values.
    
    \item \textbf{BungeeNeRF.} On the BungeeNeRF dataset, we again start from the $\lambda$ settings used in the original HAC, HAC++, and ContextGS papers for all methods, including ours. For these baselines, we use the R-D results reported by the authors at their original operating points. In addition, to obtain a PSNR range that is comparable to that of CAT-3DGS for a more reliable BD-rate computation, we further evaluate our method as well as HAC and HAC++ with one or two larger $\lambda$ values, thereby extending the PSNR variation range of these methods on BungeeNeRF.
    
    \item \textbf{PCGS.} For PCGS~\cite{chen2025pcgs}, on DeepBlending, Mip-NeRF360, and Tanks\&Temples, we train the model with $\lambda \in \{0.004, 0.008, 0.015, 0.025\}$ to enlarge the distortion range. Nevertheless, even with this wider distortion span, we still cannot obtain sufficient PSNR overlap with other methods, and we find it generally difficult to further increase the overlap for a meaningful BD-rate computation. Therefore, we include these PCGS results only for plotting the R-D curves as a qualitative comparison, and we do not include PCGS in the quantitative BD-rate evaluation. Given that expanding the overlap is consistently challenging on the above three datasets, for Synthetic-NeRF and BungeeNeRF we do not introduce new $\lambda$ values for re-training; instead, we directly use the results reported by the original authors as a reference in the R-D plots (also excluded from BD-rate). We note that our evaluation setup is largely comparable to PCGS, since both are derived from the HAC/HAC++ codebase.
\end{itemize}


\section{Limitation and Future Work}
\label{sec:limitation_future_work}
In this work, SHTC adopts transforms that is shared across all spatial regions. This global sharing keeps the parameter count and bitstream overhead small and already yields strong R–D gains over existing methods, but it also limits the ability of the transform to adapt to local statistics. A natural extension to further improve performance is to introduce limited spatial adaptivity. For example, one could partition the anchors into blocks and learn a compact bank of transforms, transmitting a few index bits to select the best transform for a given anchor or block, analogous to intra prediction mode selection in video codecs. Since SHTC is highly parameter-efficient, the additional memory footprint of such a transform bank and the overhead of signaling the transform index would remain modest. As our primary goal in this work is to demonstrate the necessity and effectiveness of incorporating transform coding into the 3DGS training process, we leave the design and evaluation of these spatially adaptive variants for future work.

\section{Supplementary Related Work}
\subsection{Unstructured Compression Paradigm}
\label{sec:add_related_work}
3DGS has recently emgered as a pioneering approach for 3D reconstruction and representation, offering both high-quality and real-time rendering. Specifically, it represents a 3D scene as a collection of Gaussian primitives. 
Each Gaussian is parameterized by a position vector and a covariance matrix, which together define its position, shape and orientation in 3D space. 
Additionally, each Gaussian is associated with an opacity parameter and a set of Spherical Harmonics (SH) coefficients to model view-dependent colors. 
Through differentiable rendering techniques, all attributes of the Gaussians can be optimized, and the number of Gaussians can be progressively increased to minimize rendering distortion. However, the unconstrained Gaussian clone/splitting operation can generate millions of Gaussians, leading to the substantial burden on storage and bandwidth. 
This motivates researchers to explore 3DGS compression. 
Three observations motivate compressing 3D Gaussian Splatting (3DGS): (i) many Gaussians have negligible impact on rendering quality; (ii) many attributes tolerate reduced numerical precision; and (iii) only a subset of regions require high-order SH coefficients for sharp view-dependent effects. Taken together, these enable effective compression via pruning, SH distillation, and quantization.

\paragraph{Pruning} The densification process of 3DGS results in an explosion in the number of Gaussians. While this allows for the reconstruction of finer scene details, a substantial portion of the Gaussians is redundant, and removing them has little effect on visual quality. The challenge lies in identifying these unimportant Gaussians. 
\cite{navaneet2024compgs} simply uses opacity as a criterion to remove transparent or nearly invisible Gaussians. \cite{ali2024elmgs,ali2024trimming} further use both gradient magnitude and opacity levels as reference to identify removable Gaussians.
Several subsequent studies assess each Gaussian's contribution to ray color prediction and utilize this as the importance score~\cite{fan2024lightgaussian,girish2024eagles,niemeyer2024radsplat}.
Lee \etal\ learn a binary mask to prune those unimportant Gaussians~\cite{lee2024compact}, which is adopted by several subsequent works~\cite{wang2024end}. \cite{hanson2025pup} propose a principled sensitivity pruning score by computing a second-order approximation of the reconstruction error with respect to the parameters of each Gaussian.
\paragraph{SH distillation} The largest portion of the memory footprint is used to store SH coefficients for modeling view-dependent color. However, a large fraction of the region consists of diffuse materials, which can be effectively modeled with view-independent color. This leads to significant waste in using high-degree SHs. To address this, \cite{papantonakis2024reducing} assign each Gaussian an appropriate SH degree to reduce waste, while \cite{fan2024lightgaussian} distill knowledge from a high-degree SH teacher model to a student model with truncated, lower-degree SHs. 
\paragraph{Scalar quantization} Gaussian attributes are typically stored at 32-bit precision. However, opacity, scale, rotation, and SH coefficients are tolerant to small inaccuracies, suggesting that lower bit-depths can represent them with minimal performance drop. In light of this, \cite{papantonakis2024reducing} propose a scalar quantization scheme that assigns each full-precision value to its nearest codeword in a K-means-derived codebook, storing only the corresponding index. 
\cite{girish2024eagles} learn a latent representation, quantize it using a uniform scalar quantizer, and employ a decoder to predict Gaussian attributes from the quantized representation. 
\paragraph{Vector quantization} Many Gaussians share similar parameters, allowing similar vectors to be mapped to a common code vector in a codebook. Instead of storing high-precision vectors, only compact indices are needed. Based on this insight, several works apply vector quantization to eliminate the redundancy within Gaussian attributes~\cite{navaneet2024compgs,fan2024lightgaussian,niedermayr2024compressed}. Furthermore, \cite{wang2024end} employ entropy-constrained vector quantization, incorporating entropy coding into the quantization process. It assigns shorter codes to those code vectors that are used more frequently, effectively compressing the index stream.

\subsection{Post-training Transform Coding Paradigm}
\label{sec:3dgs_post_training_transform}
In this subsection, we focus on methods that employ transforms during the post-training stage of 3DGS compression. 

Applying transform coding to 3DGS is challenging because 3DGS data are irregular and unordered. To sidestep these challenges, prior work defers the application of transform coding until the post-training stage, once the 3DGS geometry is fixed. 
Two main strategies have been explored. First, some methods project 3DGS onto a set of feature planes or adopt a triplane representation~\cite{chan2022efficient} to implicitly encode Gaussian attributes, then compress the learned planes with standard image or video codecs~\cite{lee2025compression, morgenstern2024compact}. Second, other methods treat a pretrained 3DGS model as a point cloud and apply point-cloud attribute transforms such as Graph Fourier transform (GFT)~\cite{sandryhaila2013discrete} and Region-adaptive Hierarchical Transform (RAHT)~\cite{de2016compression} to compress the model~\cite{xie2024mesongs,yang2024benchmark,huang2025hierarchical,wang2025adaptive,yang2025hybridgs}. The above methods borrow transforms from images, video, or point clouds but apply fixed, linear transforms only after training. This decoupling blocks mutual adaptation between the transform and the 3DGS representation, often limiting R–D performance. In contrast, we propose an end-to-end transform-coding framework that jointly adapts the transform, entropy model and the 3DGS representation, significantly improving compression performance over post-training transform approaches.

\end{document}